\documentclass[final,5p,times,twocolumn]{elsarticle}

\usepackage{amssymb}

\usepackage{booktabs}
\usepackage{xcolor}

\definecolor{cred}{HTML}{FF0000}
\definecolor{cdarkgreen}{HTML}{009051}
\definecolor{cdarkyellow}{HTML}{CCCC00}

\usepackage{amsmath, amssymb}

\DeclareMathOperator*{\argmin}{\arg\!\min}

\usepackage{xfrac}

\usepackage{algorithm}
\usepackage[noend]{algorithmic}

\makeatletter
\renewcommand{\ALG@name}{Algo.}
\makeatother

\makeatletter
\newcommand{\algorithmicbreak}{\textbf{break}}
\newcommand{\BREAK}{\STATE \algorithmicbreak}
\makeatother

\usepackage{subfig}

\usepackage{rotating}

\usepackage{scalerel}
\usepackage{tikz}
\usetikzlibrary{svg.path}

\usepackage{url}

\journal{arXiv.org}

\begin{document}

\begin{frontmatter}



\title{Implicit Shape Model Trees: Recognition of 3-D Indoor Scenes\\and Prediction of Object Poses for Mobile Robots}

\tnotetext[t1]{This article is accompanied by six supplementary video clips showing demonstrations of scenes by humans and experiments with a physical mobile robot. This work was supported by the ' DFG - German Research Foundation' [Grant 255319423]. Source code for the presented Active Scene Recognition system is publicly available at \url{http://wiki.ros.org/asr}.}

\author[label1]{Pascal Meißner\corref{cor1}}
\ead{pascal.meissner@abdn.co.uk}

\author[label2]{Rüdiger Dillmann}
\ead{dillmann@kit.edu}

\cortext[cor1]{Corresponding author}

\affiliation[label1]{organization={School of Engineering, University of Aberdeen, Scotland},
            country={United Kingdom}}

          
\affiliation[label2]{organization={Humanoids and Intelligence Systems Lab, Institute for Anthropomatics and Robotics, Karlsruhe Institute of Technology},
            country={Germany}}



\begin{abstract}
We present an approach for mobile robots to recognize scenes in object arrangements distributed across cluttered environments. Recognition is enabled by intertwining the robot's search for objects and the assignment of found objects to scenes. Our scene model called "Implicit Shape Model (ISM) trees" allows these two tasks to be solved jointly. This article presents novel algorithms for ISM trees to recognize scenes and predict poses of searched objects. We define scenes as object sets in which some objects are connected via 3-D spatial relations. In previous work, we recognized scenes with single ISMs. However, single ISMs are prone to false positives. As a remedy, we have developed ISM trees, a hierarchical model consisting of multiple ISMs. This article contributes a recognition algorithm that now enables the use of ISM trees for scene recognition. ISM trees should be ideally generated from human demonstrations of object arrangements. As a suitable algorithm was not available, we introduce such a generation algorithm. In line with the active vision paradigm, we combined scene recognition and object search in previous work. However, an efficient algorithm was lacking to make this combination effective. Physical experiments show that this is now overcome with a new algorithm achieving efficient combination through predicted object poses.
\end{abstract}


\end{frontmatter}


\section{Introduction}\label{sec:introduction}

To act autonomously in various situations, robots not only need capabilities to perceive and act but must also be provided with models of the possible states of the world. If we imagine such a robot as a household helper, it will have to master tasks such as setting, clearing, or rearranging tables. Let us imagine that such a robot looks at the table in \figurename~\ref{fig:int_intro} and tries to determine which of these tasks is pending. More precisely, the robot must choose between four different actions, each of which contributes to the solution of one of the tasks. An autonomous robot may choose an action based on a comparison of its perceptions with its world model, i.e., its assessment of the state of the world. Which scenes are present is an elementary aspect of such a world state. Modeling scenes and comparing their models with perceptions is the topic of this article. In particular, we model scenes not by the absolute poses of the objects in them, but by the spatial relations between these objects. Such a model can be more easily reused across different environments because it models a scene regardless of where it occurs.

\subsection{Scene Recognition --- Problem and Approach}

\begin{figure}[tpb]
  \centering
      \includegraphics[width=1\linewidth]{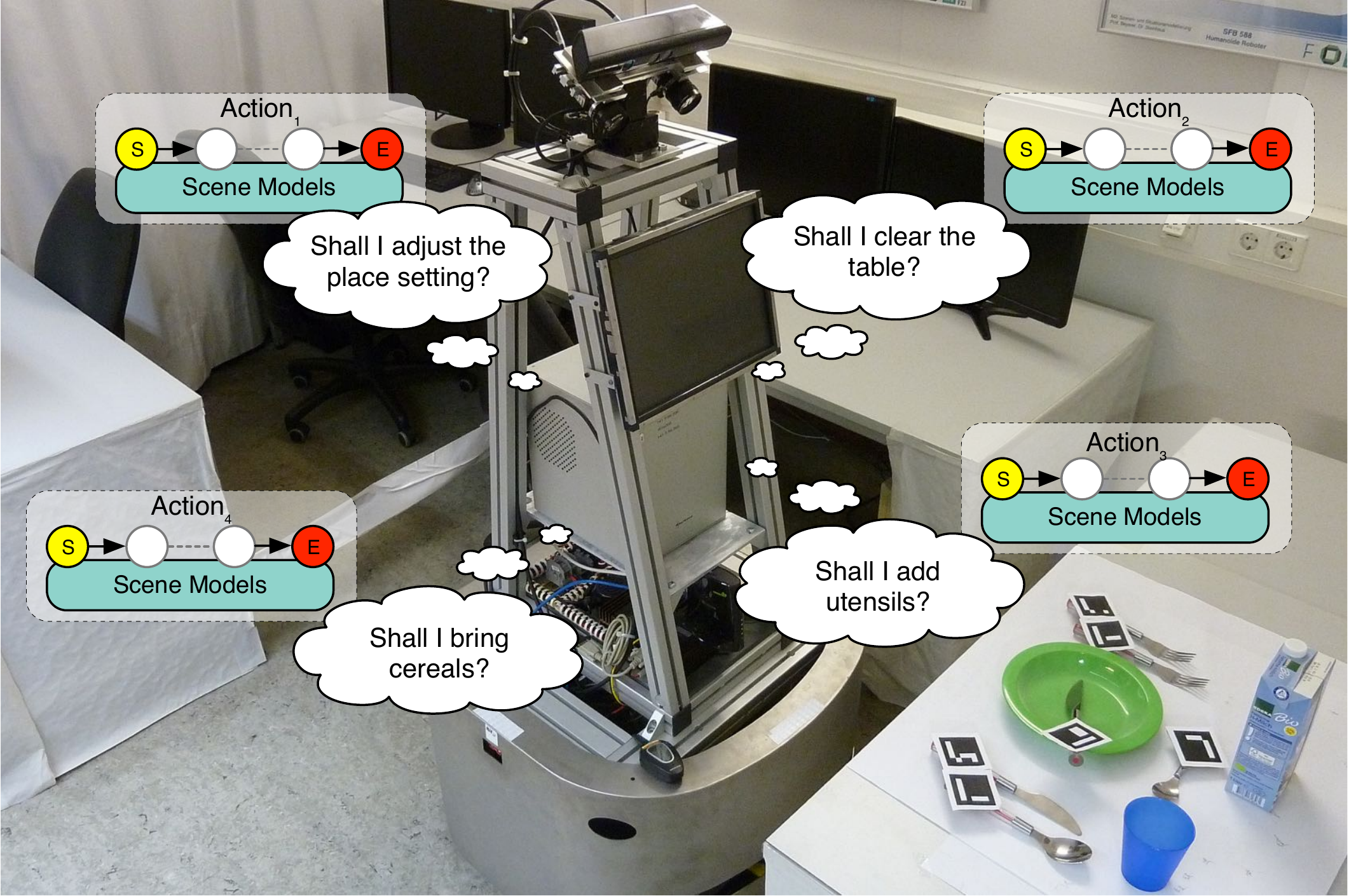}
  \caption{Motivating example for scene recognition: Mobile robot looks at an object configuration (arrangement). It reasons which of its actions to apply.}
  \label{fig:int_intro}
\end{figure}

This article provides a solution to the problem of classifying into scenes a configuration or so-called "arrangement" of objects whose poses are given in six degrees of freedom (6-DoF). Recognizing scenes based on the objects present is an approach suggested for indoor environments by \cite{quattoni2009recognizing} and successfully investigated by \cite{espinace2010indoor}. The classifier we propose not only describes a single configuration of objects, but rather a multitude of configurations that these objects can take on while still representing the same scene. Hereinafter, this multitude of configurations will be referred to as a ``scene category'' rather than as ``scene'' which is a specific object configuration. For \figurename~\ref{fig:int_intro}, the classification problem we address can be paraphrased as follows: ``Is the present tableware an example of the modeled scene category?", ``How well does each of these objects fit our scene category model?", ``Which objects on the table belong to the scene category?", and ``How many objects are missing from the scene category?". Our classifier is learned from object configurations demonstrated by a human in front of a robot and perceived by the robot using 6-DoF object pose estimation.

Many scene categories require that the spatial characteristics of relations, including uncertainties, be accurately described. For example, a table setting requires that some utensils be exactly parallel to each other, whereas their positions relative to the table are less critical. To meet such requirements, we proposed single Implicit Shape Models (ISMs) as scene classifiers in \cite{meissner2013}. Inspired by Hough voting, a classic but still popular approach (see \cite{qi2019deep}, \cite{sommer2020primitect}), our ISMs let each detected object in a configuration vote on which scenes it might belong to, using the spatial relations in which the object participates. The fit of these votes yields a confidence level for the presence of a scene category in an object configuration. Overall, this article is not about feature extraction, but about modeling relations and their variations in 6-DoF. ISMs for scene recognition should not be considered as an alternative but as a complement to the immensely successful Convolutional Neural Nets (CNNs).

\begin{figure*}[tpb]
  \centering
      \includegraphics[width=1\linewidth]{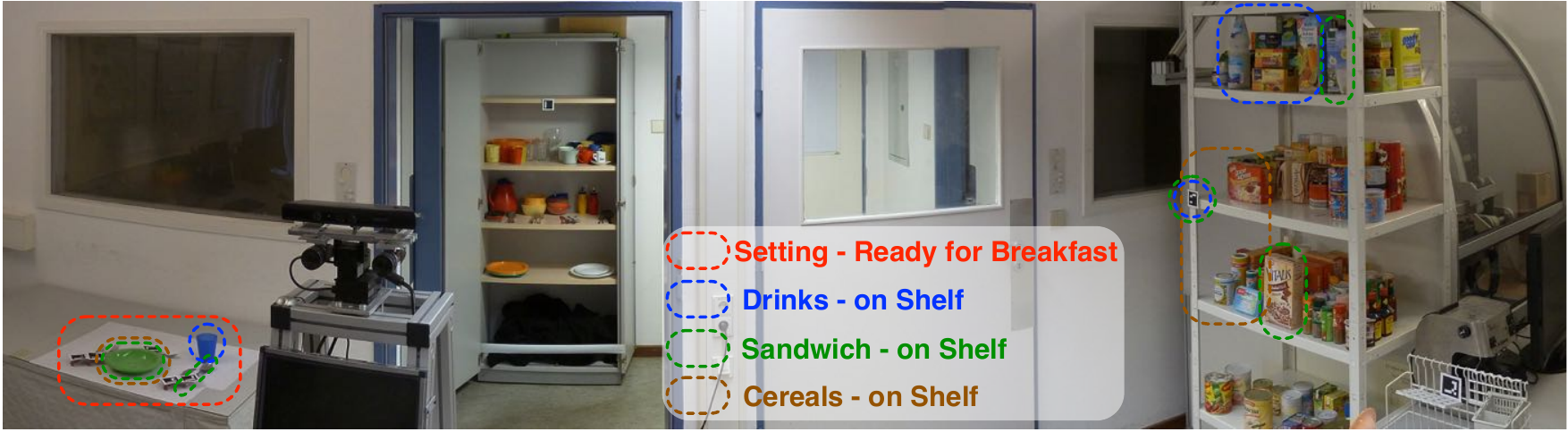}
      \caption{Experimental setup mimicking a kitchen. The objects are distributed over a table, a cupboard, and some shelves. Colored dashed boxes are used to discern the searched objects from the clutter and to assign objects to exemplary scene categories.}
  \label{fig:int_pano}
\end{figure*}

\subsection{Object Search --- Problem and Approach}

\figurename~\ref{fig:int_pano} shows our experimental kitchen setup as an example of the many indoor environments where objects are spatially distributed and surrounded by clutter. A robot will have to acquire several points of view before it has observed all objects in such a scene. This problem is addressed in a field called three-dimensional object search (\cite{ye1999sensor}). The existing approaches often rely on informed search. This method is based on the fact that detected objects specify areas where other objects should be searched. However, these areas are predicted by individual objects rather than by entire scenes. Predicting poses utilizing individual objects can lead to ambiguities, since, e.g., a knife in a table setting would expect the plate to be beneath itself when a meal is finished, whereas it would expect the plate to be beside itself when the meal has not yet started. Using, instead, estimates for scenes to predict poses resolves this problem.

To this end, we presented 'Active Scene Recognition' (ASR) in \cite{meissner2014} and \cite{meissner2016}, a procedure that integrates scene recognition and object search. Roughly, the procedure is as follows: The robot first detects some objects and computes which scene categories these objects may belong to. Assuming these scene estimates are valid, it then predicts where missing objects, which would also belong to these estimates, could be located. Based on these predictions, camera views are computed for the robot to check. In this article, ASR is detailed in Sec. \ref{sec:asr}. In Sec. \ref{sec:ear_eoasr}, we evaluate ASR and this article's contributions to it on a physical robot. To distinguish between ASR and pure scene recognition, the latter is referred to as 'Passive Scene Recognition' (PSR). PSR is detailed in Sec. \ref{sec:psr}. The flow of our overall approach (see \cite{meissner2018}), which consists of two phases: first the learning of scene classifiers and then the execution of Active Scene Recognition, is shown in \figurename~\ref{fig:int_pbd_cycles_new}.

\subsection{Relation Topology Selection --- Problem}

We train our scene classifiers from sensory-perceived demonstrations (see \cite{kroemer2019review}), which consist of a two- to low-three-digit number of recorded object configurations. This learning task involves the problem of selecting pairs of objects in a scene to be connected by spatial relations. Which combination of relations is modeled determines the number of false positives returned by a classifier and the runtime of scene recognition. Combinations of relations are hereinafter referred to as relation topologies. Whereas such topologies contain only binary relations, they can represent many ternary or n-ary relations with multiple binary ones. In Sec. \ref{sec:rts_co}, we motivate and outline how we selected relation topologies in our previous work in \cite{meissner2015}.

\subsection{Contributions and Differences from Previous Work}

To not unnecessarily restrict this topology selection, a scene classifier should be able to represent a maximum of topologies. To this end, we first suggested 'Implicit Shape Model trees', a hierarchical scene model, in \cite{meissner2013}. This model consists of multiple ISMs stacked upon each other, with the ISMs in it representing different portions of the same scene category. Such portions are brought together by additional ISMs in such a tree. The closer an ISM is to the root of a tree, the larger the portion it covers. However, while we outlined such a tree in \cite{meissner2013}, we did not define how scene recognition in terms of data or control flow would work with ISM trees. In Sec. \ref{sec:psr_rwismt}, we close this gap which prevents greater use of ISM trees by contributing an algorithm for scene recognition with ISM trees. In \cite{meissner2015}, we defined how to select relation topologies. However, we did not describe how ISM trees are generated from such topologies. In Sec. \ref{sec:psr_loismt}, we contribute an algorithm for generating ISM trees, thus closing another essential gap.

\begin{figure}[tpb]
  \centering
      \includegraphics[width=1\linewidth]{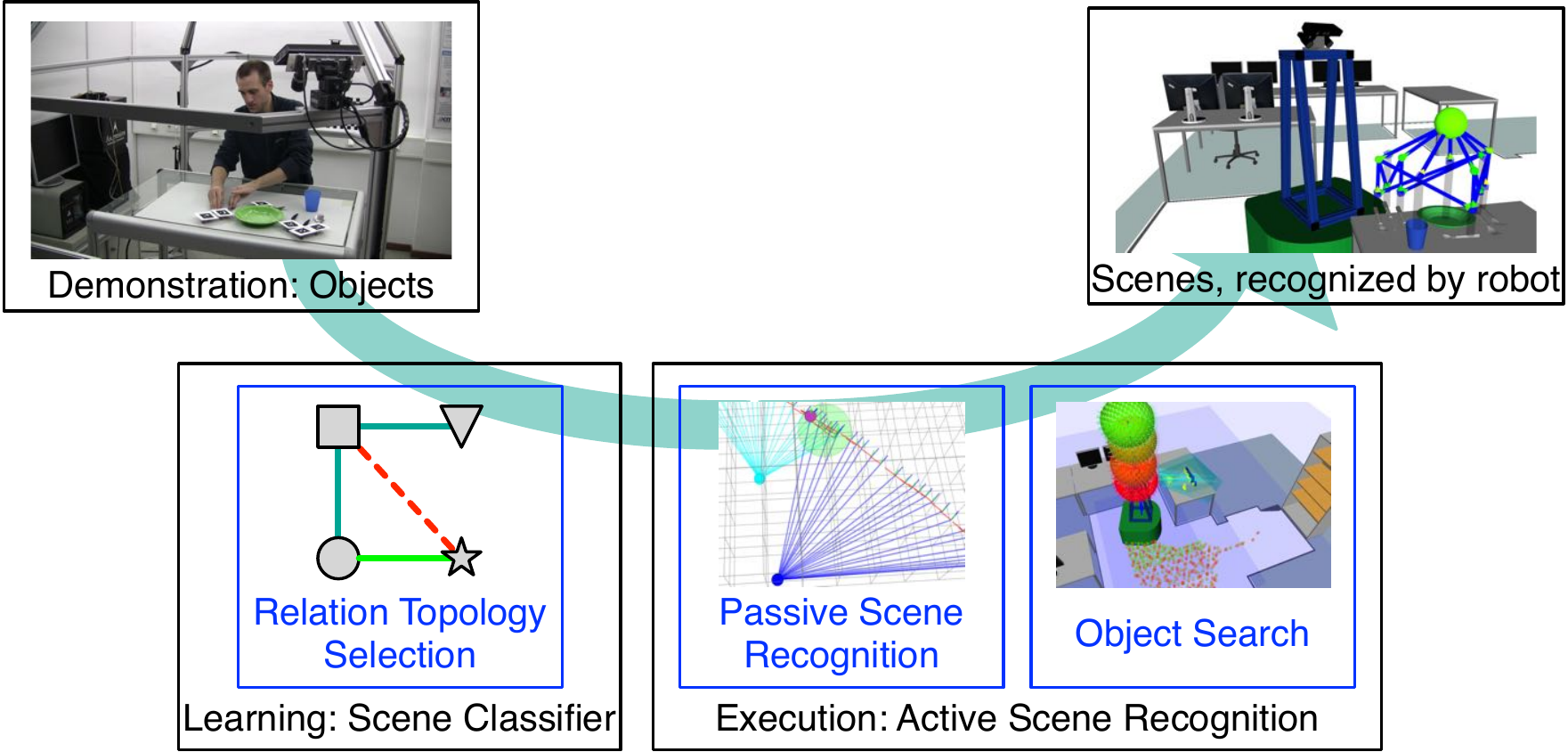}
      \caption{Overview of the research problems (in blue) addressed by our overall approach.  At the top are the inputs and outputs of our approach. Below are the two phases of our approach: Scene classifier learning and ASR execution.}
  \label{fig:int_pbd_cycles_new}
\end{figure}

As visible in \figurename~\ref{fig:int_pbd_cycles_new}, learned ISM trees are used to perform ASR. To make ASR possible, we had to link two research problems: Scene recognition and object search. To this end, we proposed a technique for predicting the poses of searched objects in \cite{meissner2014} and reused it in \cite{meissner2016}. However, this technique suffered from a combinatorial explosion. We close this gap, which made ASR impractical for larger scenes, by contributing a prediction algorithm in Sec. \ref{sec:asr_popomo} that efficiently predicts object poses. In summary, this work’s \textbf{contributions} are an:

\begin{enumerate}
\item Algorithm for generating ISM trees
\item Algorithm for scene recognition using ISM trees
\item Algorithm for predicting object poses with ISM trees
\end{enumerate}

\subsection{Equipment and Constraints}

For the experiments with our robot MILD in \figurename~\ref{fig:int_intro}, we integrated these algorithms into ASR. The robot consists of a mobile base and a pivoting camera head. Searched objects are detected using third-party object pose estimators. From all searched objects in our kitchen setup, only the utensils are localized using markers. ISM trees provide two parameters that set the degree to which object poses may deviate from the modeled relations without being excluded from the scene. These parameters were tested in the range of [mm] to [dm] for object positions and in the one- to two-digit [°] range for object orientations. Because ISM trees emphasize the modeling of relations, they focus on the objects in a scene. They complement work as \cite{huang2020indoor} that place emphasis on the global shape of a scene including the walls or the floor. ASR can only search objects that are part of demonstrated scene categories. ASR also assumes that the environment is static during object search, as opposed to approaches (\cite{burschka2019spatiotemporal}) that address dynamic scenes.

\section{Related Work}\label{sec:relatedwork}

\subsection{Scene Recognition}

Scene understanding is generally understood as an image labeling problem. It can be addressed by two approaches. Either scenes are derived from detected objects and relations between them, or scenes are derived directly from image data without concepts such as objects, as by \cite{zhou2014learning}. Descriptions of scenes in the form of graphs (including existing objects and relation types), which are derived by an object-based approach, are far more informative for further use, as in our work for object search, than the global labels for images that are instead derived using the "direct" approach. Work by \cite{xu2017scene} or \cite{zellers2018neural} that follows the object-based approach relies on neural nets for object detection including feature extraction (e.g. by \cite{ren2015faster}, \cite{redmon2016you}), which they combine with neural nets to generate scene graphs. This is enabled by datasets that include relations (\cite{krishna2016visual}, \cite{kuznetsova2018open}), which have been published in recent years alongside object detection datasets (\cite{deng2009imagenet}, \cite{lin2014microsoft}). These scene graph nets are very powerful but are designed with the goal of learning models of relations that focus on relation types or meanings rather than the spatial characteristics of relations. In contrast, in our work, we want to focus on accurately modeling the spatial properties of relations and their uncertainties. Yet, our model should be able to cope with small amounts of data, since we want to learn it from demonstrations of people's personal preferences concerning object configurations. Indeed, personal data must be provided by a user, wherein users wants to put a limited effort.

Examples for preferences in object configurations can be breakfast tables which hardly two people will want to have set in the same way. Yet, people will expect household robots to take their preferences into account when arranging objects. For example, \cite{abdo2016organizing} address personal preferences by combining a relation model for learning preferences for arranging objects on a shelf with object detection. However, while their approach can even successfully handle conflicting preferences, it can also miss subtle differences between spatial relations and is therefore too coarse for us. Classifiers explicitly designed to model relations and their uncertainties such as the part-based models \cite{grauman2011visual} from the 2000s are a more expressive alternative. They also have low sample complexity, making them suitable for learning from demonstrations. Replacing their outdated feature extraction component with CNN-based object detectors or pose estimators (e.g. DOPE \cite{tremblay2018deep} or PoseCNN \cite{xiang2017posecnn}), we obtain an object-based scene classifier that combines the power of CNNs with the expressiveness of part-based models in relation modeling. Thus, in our approach, we combine pre-trained object pose estimators with the relation modeling of part-based models.

Ignoring the outdated feature extraction of part-based models, we note that \cite{ranganathan2007semantic} already successfully used a part-based model, the Constellation Model \cite{fergus2003object}, to represent scenes. Constellation models define spatial relations using a parametric representation over cartesian coordinates (a normal distribution), just like the Pictorial Structures Models \cite{felzenszwalb2005pictorial} (another part-based model) do. Recently, \cite{Kartmann2020}'s approach of using probability distributions over polar coordinates to define relations is proving more effective for describing practically relevant relations. Whereas such distributions are more expressive than \cite{abdo2016organizing}'s model, they are still too coarse for us. Moreover, they use a fixed number of parameters to represent relations. What would be most appropriate when learning from demonstrations of varying length is a relation model whose complexity grows with the number of training samples demonstrated, i.e., a non-parametric model \cite{kroemer2019review}. One such flexible models are the Implicit Shape Models (ISMs) of \cite{leibe2004combined}, \cite{leibe2008robust}. Therefore, we chose ISMs as the basis for our approach. One shortcoming that ISMs have in common with Constellation and Pictorial Structures Models is that they can only represent a single type of relation topology. However, the topology that yields the best tradeoff between the number of false positives and scene recognition runtime can vary from scene to scene. To account for this, we extended the ISMs to our hierarchical ISM trees. We also want to mention scene grammars (\cite{yubottom2020}) which are similar to part-based models but motivated by formal languages. They again model relations probabilistically and use only star topologies. For these reasons, we prefered ISMs over scene grammars.

\subsection{Object Pose Prediction}

The search of objects in 3-D has been addressed as an active-vision (\cite{shubina2010visual}, \cite{aydemir2011search}, \cite{eidenberger2012scene}, \cite{rasouli2019attention}) and a manipulation problem (\cite{wong2013manipulation}, \cite{dogar2014object}, \cite{li2016act}). Active-vision approaches are divided into direct (\cite{druon2020visual}, \cite{hernandezefficient2020}) and indirect (\cite{garvey1976perceptual}) search depending on the type of knowledge about potential object poses used. Indirect search uses spatial relations to predict from known poses of objects those of searched objects. Indirect search can be classified according to the type (\cite{thippur2015comparison}) of relations used to predict poses. \cite{Southey}, \cite{lorbach2014prior}, \cite{zeng2020semantic} are examples of using relations that correspond to natural language concepts such as 'above' in Robotics. Even though such symbolic relations provide high-quality generalization, they can only provide coarse estimates of metric object poses - too coarse for many object search tasks. For example, \cite{bozcan2019cosmo} successfully adapted and used Boltzmann Machines to encode symbolic relations. Representing relations metrically with probability distributions showed promising results in \cite{kunze2014using}. However, their pose predictions are derived exclusively from known locations of individual objects, which leads to ambiguities that can be avoided when using scenes instead.

\begin{figure}[tpb]
  \centering
      \includegraphics[width=1\linewidth]{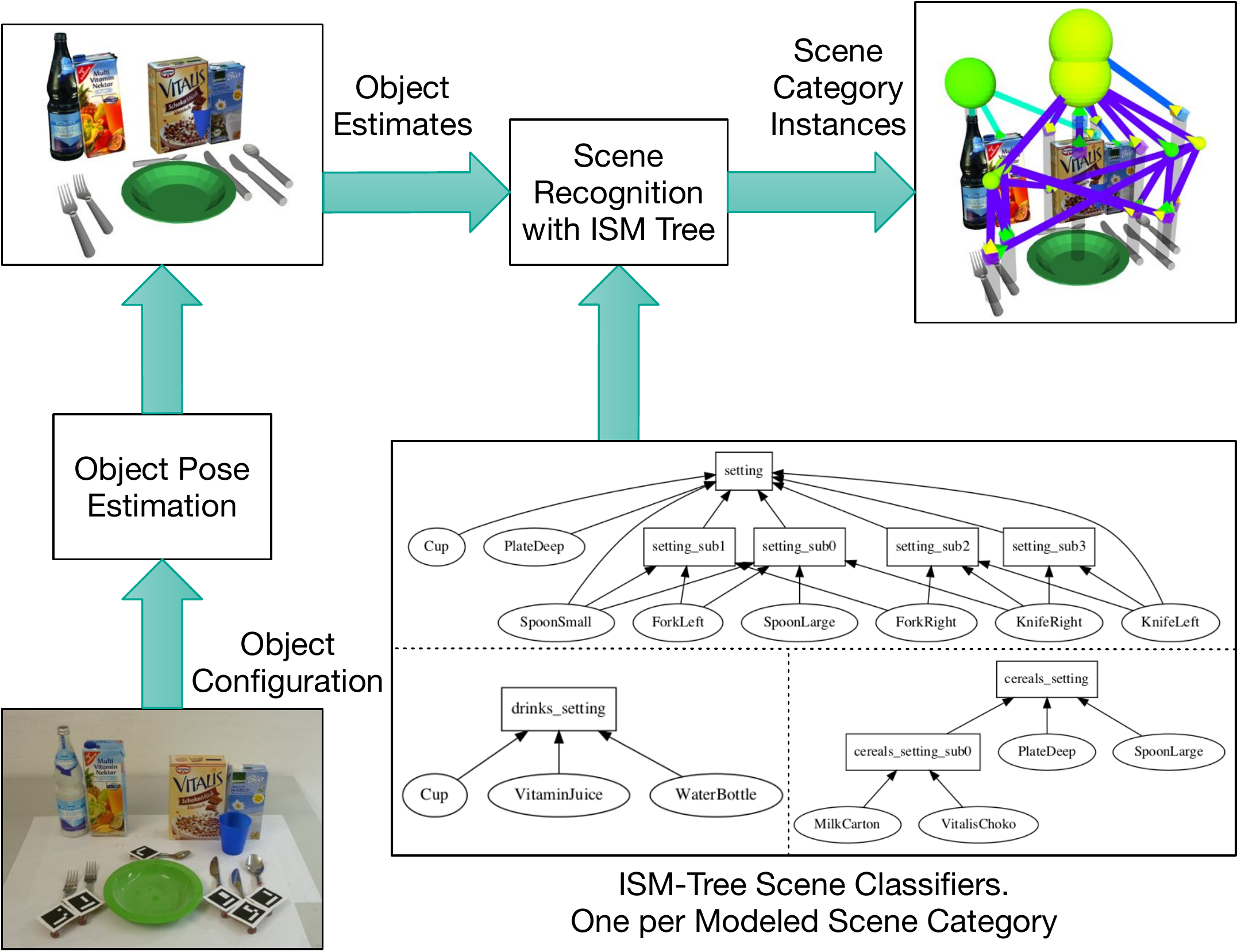}
  \caption{Overview of the inputs and outputs of \textbf{Passive Scene Recognition}. Spheres represent increasing confidences of outputs by colors from red to green.}
  \label{fig:int_scene_recognition_pheno}
\end{figure}


\section{Methods --- Passive Scene Recognition}\label{sec:psr}

\begin{figure*}[tpb]
  \centering
      \includegraphics[width=1.0\linewidth]{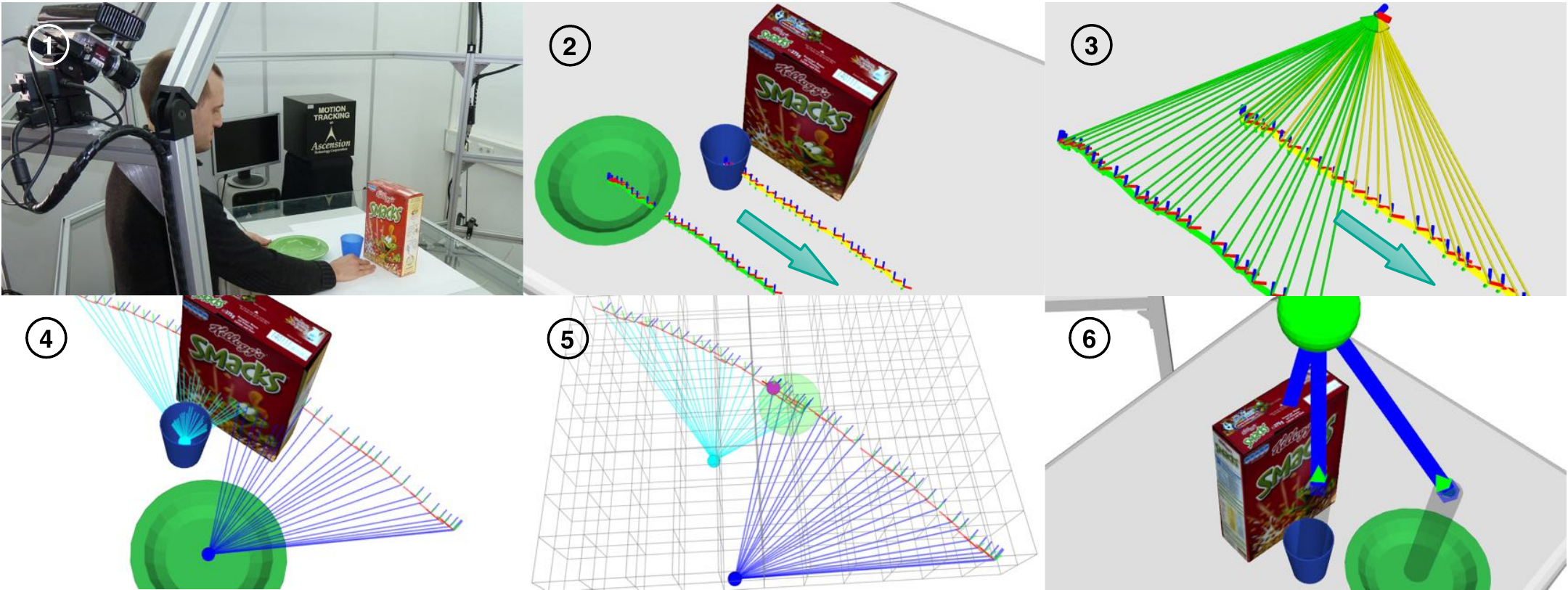}
  \caption{\textbf{1}: Snapshot of a demonstration. \textbf{2}: Demonstrated object trajectories as sequences of object estimates. The cup is always to the right of the plate, and both in front of the box. \textbf{3}: Relative poses, visualized as arrows pointing to a reference, form the relations in our scene classifier. Here the classifier models the relations "cup-box" and "plate-box". \textbf{4}: Objects voting on poses of the reference, using the relations from \textbf{3}. \textbf{5}: Accumulator filled with votes. \textbf{6}: Here the cup is to the left of the plate, which contradicts the demonstration. The learned ISM does not recognize this, as it does not model the "cup-plate" relation. It outputs a false positive.}
  \label{fig:psr_fruehstueck}
\end{figure*}

\subsection{Overview and ISMs as by Leibe et al. (\cite{leibe2004combined}, \cite{leibe2008robust})}\label{sec:psr_co_1}

In Sec. \ref{sec:introduction}, we introduced scene recognition as a black box that receives estimates for objects as input. From these, it derives as output the present instances of scene categories. The concrete process of how our approach to scene recognition works is shown in \figurename~\ref{fig:int_scene_recognition_pheno}. Firstly, external object pose estimators derive the types and poses of the present objects. Hence, a physical object configuration is transformed into a set of 'object estimates'. These estimates are passed on to each of our scene classifiers. Every scene classifier then returns estimates for the presence of instances\footnote{The confidence level of such an instance is visualized by a sphere above the scene. Its color changes from red to green when the confidence increases. Relations are shown as lines whose colors indicate to which ISM they belong.} of one scene category, including where these instances are located in 3-D space. Each scene classifier models a scene category and is learned from a recording of a human demonstration. Such a demonstration of object configurations is shown in \textbf{2} in \figurename~\ref{fig:psr_fruehstueck}. A plate and a cup are pushed from left to right, yielding two parallel trajectories\footnote{Trajectories are visualized as line strips between coordinate frames that stand for object poses.}. Please note that we use pre-trained object pose estimators to record object poses during demonstrations. One of the demonstrated configurations is visible in \textbf{1} in \figurename~\ref{fig:psr_fruehstueck}.

In \cite{meissner2013}, we redefined Implicit Shape Models (ISMs) so they would represent scenes (which consist of objects), instead of objects (which consist of object parts). The original ISMs were used to recognize objects in 2-D images and similar to the Generalized Hough Transform (\cite{ballard1981generalizing}). As \cite{nixon2012feature} write, this transform is used to ``locate arbitrary shapes with unknown position, size and orientation'' in images. It does so by gathering evidence about the said properties of a shape from the pixel values in an image. Evidence gathering is implemented by learning a mapping from pixels to shape parameters for each shape. Using this mapping, pixels cast votes in an accumulator array for different parameter values. The parameter values of the present shapes can then be determined from the local maxima in this array.

\subsection{Single ISMs as Scene Classifiers --- Previous Work}\label{sec:psr_co_2}

The learning (\cite{meissner2013}) of a similar mapping for our 'scene classifier ISMs' can be thought of as adding entries to a table, similar to K-Nearest Neighbors (\cite{mitchell1997machine}). In \cite{meissner2013}, we transformed the absolute poses in the trajectories of interrelated objects into relative poses which we then stored in a table. Thus, the learning did not involve the optimization of parameters of a model. Instead, ISMs represented spatial relations as sequences of 6-DoF relative poses. Nevertheless, scene recognition could be done efficiently as it mainly consisted of highly parallelizable matrix operations. We did not arbitrarily decide which pairs of absolute poses to convert into relative poses. Instead, among the objects in a scene category, we did set one as the reference object to which all relative poses pointed. In \textbf{2} in \figurename~\ref{fig:psr_fruehstueck}, for example, the cornflakes box is the reference. Accordingly, all relative poses in \textbf{3} in \figurename~\ref{fig:psr_fruehstueck}\footnote{Each relation consists of all visualized arrows of one color.} point from the plate and the cup to the box.

Like the Generalized Hough Transform, scene recognition (\cite{meissner2013}) with a single ISM started with a vote. Instead of letting pixels vote, the known objects cast votes starting from the place where they had been located. The voting was done by combining the estimated pose of the respective object with all those relative poses in the table of the ISM, assigned to this object. The visualization of a vote in \textbf{4} in \figurename~\ref{fig:psr_fruehstueck} shows at which poses the plate and the cup respectively expect the box. The votes cast are entered into the 3-D accumulator array shown in \textbf{5} in \figurename~\ref{fig:psr_fruehstueck}. Once the voting was completed, we searched this array for the most comprehensive and consistent combinations of votes from the objects in a scene category. We identified the top-rated combinations as instances of the scene category the ISM modeled. Note that it did not matter whether the reference object was present in that combination and that a missing reference did not cause recognition to fail.

To avoid a combinatorial explosion during this search, we only compared votes that had fallen into the same bin of the accumulator, using a method similar to the Mean-Shift Search proposed by \cite{leibe2008robust}. This procedure allowed for discarding votes from irrelevant objects, i.e., objects that did not belong to the modeled scene category. Since a single ISM only ever relates one reference object to all other objects in a scene, it can only represent a star-shaped topology of relations. This could lead to false positives in scene recognition, as long as only relations between the other objects would be violated. For example, in \textbf{6} in \figurename~\ref{fig:psr_fruehstueck}, we swapped the cup and the plate. Nevertheless, the ISM considers this configuration a valid instance of its scene category. Hence, star-shaped topologies and single ISMs are not sufficient to reliably recognize many scene categories.

\subsection{Implicit Shape Model Trees --- Outline}\label{sec:psr_co_3}

Instead, we create a scene classifier that supports all connected relation topologies by first partitioning the given relation topology into star-shaped subtopologies, which are then assigned to separate ISMs. Based on this partitioning, we assemble the ISMs and connect them into a tree, creating a compound hierarchical model of the initial relation topology: The ISM tree. To avoid a combinatorial explosion when using an ISM tree for scene recognition, we take the precaution that only a restricted amount of data, the most comprehensive and consistent combinations of votes in each ISM, is shared between connected ISMs. Such an approach could have led to false negatives in scene recognition. However, such an effect is not observed during our experiments in Sec. \ref{sec:ear}. Before we detail this article's contributions to the ISM trees, we present the assumptions and notation used throughout the article in Sec. \ref{sec:psr_srd} and outline a technique from our previous work which partitions connected topologies into stars in Sec. \ref{sec:psr_rtp}. Sec. \ref{sec:psr_loismt} introduces a novel algorithm for generating ISM trees from these stars. Yet another contribution, we present an algorithm for recognizing scenes with ISM trees in Sec. \ref{sec:psr_rwismt}.

\begin{figure*}[tpb]
  \centering
      \includegraphics[width=1.0\linewidth]{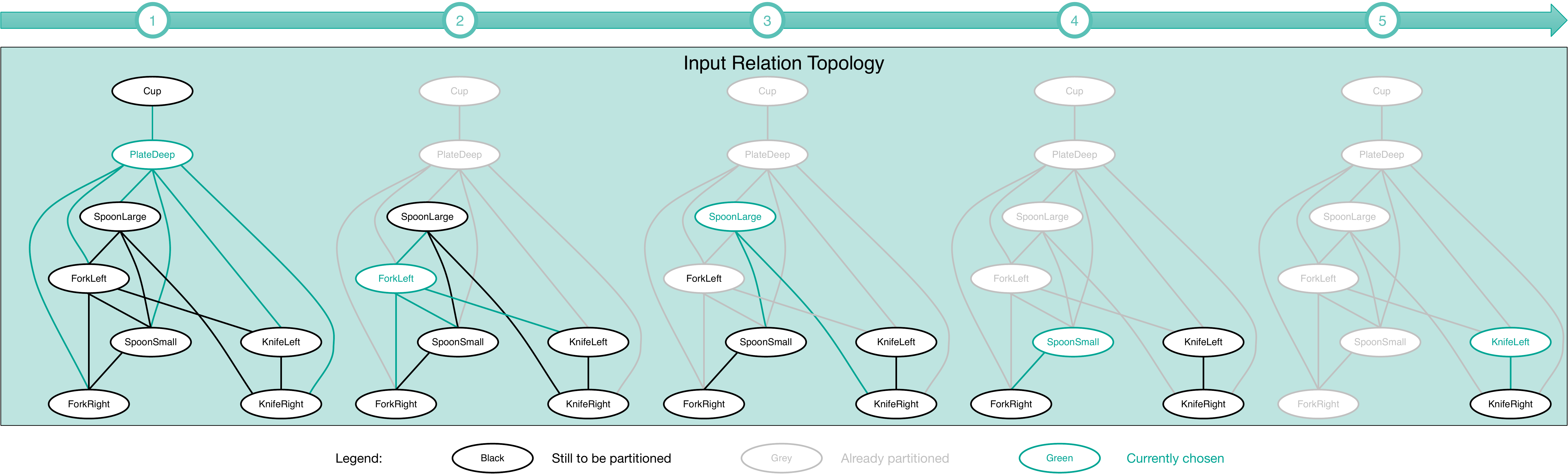}
      \caption{Algorithm - How the connected relation topology - from which we generate the ISM tree in \figurename~\ref{fig:psr_ism_tree_recog_interrel} - is first partitioned. The partitioning includes five iterations. It starts at the leftmost graph and ends at the rightmost. In each iteration, a portion of the connected topology, colored green, is converted into a separate star-shaped topology. All stars are on the left of \figurename~\ref{fig:psr_learn_ism_tree}.}
  \label{fig:psr_factor_out}
\end{figure*}

\subsection{Preliminaries --- Definitions for Scene Recognition}\label{sec:psr_srd}

We define an object \(o\) as an entity whose state \(\textbf{E}(o,t) = (c, d, \textbf{T})\) at a point in time \(t\) is estimated from sensor data. The state is described by a triple consisting of a label \(c\) indicating the object class, a label \(d\) used to distinguish between different objects of the same class, and a transformation matrix \(\textbf{T} \in \mathbb{R}^{4\times4}\) indicating the pose of the object. A scene category \(\textbf{S} = (\{o\},\{\textbf{R}\})\) consists of objects and the spatial relations \(\{\textbf{R}\}\) between the objects. The identity of a scene category is defined by a label \(z\) and each spatial relation is represented as a set of relative 6-DoF poses \(\{\textbf{T}_{jk}\}\). In scene recognition, the fit between a configuration \(\{\textbf{E}(o,t)\}\) of objects (a set of states) and the model of a scene category is estimated. If this fit, whose degree is indicated by a confidence level \(b(\textbf{I}_{\textbf{S}}) \in [0,1]\), is sufficiently good, we consider the objects as an instance \(\textbf{I}_{\textbf{S}}\) of the scene category and locate it at a pose \(\textbf{T}_{F}\). Models of scene categories are learned from trajectories demonstrated over \(l\) time steps for each object included in the category. Each trajectory is a sequence \(\textbf{J}(o) = (\textbf{E}(o, 1),\dots, \textbf{E}(o, l))\) of estimates of the time-variant state \(\textbf{E}(o,t)\) of an object.

When modeling a scene category with an ISM tree, pairs of trajectories are converted into spatial relations. The relations are stored in a table, as outlined in Sec. \ref{sec:psr_co_2}. A relation topology \(\Sigma = (\{o\},\{\textbf{R}\})\) describes the same as a scene category, but at a different level of abstraction. In a topology, relations are represented on a purely algebraic level instead of explicitly considering their spatial properties as scene categories do. We distinguish the following types of topologies: Star topologies \(\Sigma_{\sigma}\), in which a single object \(o_{F}\) (the reference object) is connected to all other objects by one relation each. Complete topologies, in which every object is connected to all other objects. Connected topologies \(\Sigma_{\nu}\), in which each pair of objects is connected by a sequence of relations.

\subsection{Relation Topology Partitioning --- Previous Work}\label{sec:psr_rtp}

An ISM tree is learned in two steps. In step one, the relations in a scene category are distributed across several ISMs. Step one is covered in this subsection and is part of our previous work (\cite{meissner2015}). In step two, the ISMs are then combined into a hierarchical scene classifier. We refer to step two as the tree generation which is one of the three contributions of this article. It is introduced in the next subsection. Let us assume for step one that a connected relation topology \(\Sigma_{\nu}\) was given for a scene category \(\textbf{S}\). Step one distributed the relations in the scene category by partitioning this so-called input topology \(\Sigma_{\nu}\) into a set of star-shaped subtopologies \(\{\Sigma_{\sigma}(j)\}\). The partitioning was performed using a depth-first search that successively selected objects \(o_{M}\) in the topology that were involved in as many relations \(\textbf{R}\) as possible. We considered each selected object as the center of a star topology \(\Sigma_{\sigma}(j) = (o_{M} \cup N(o_{M}), \{\textbf{R}_{M}\})\). This star topology also included the relations \(\{\textbf{R}_{M}\}\) in which the center participated and the neighborhood \(N(o_{M})\) of the center, i.e., all objects connected to the center by the relations \(\{\textbf{R}_{M}\}\).

\begin{figure*}[tpb]
  \centering
      \includegraphics[width=1\linewidth]{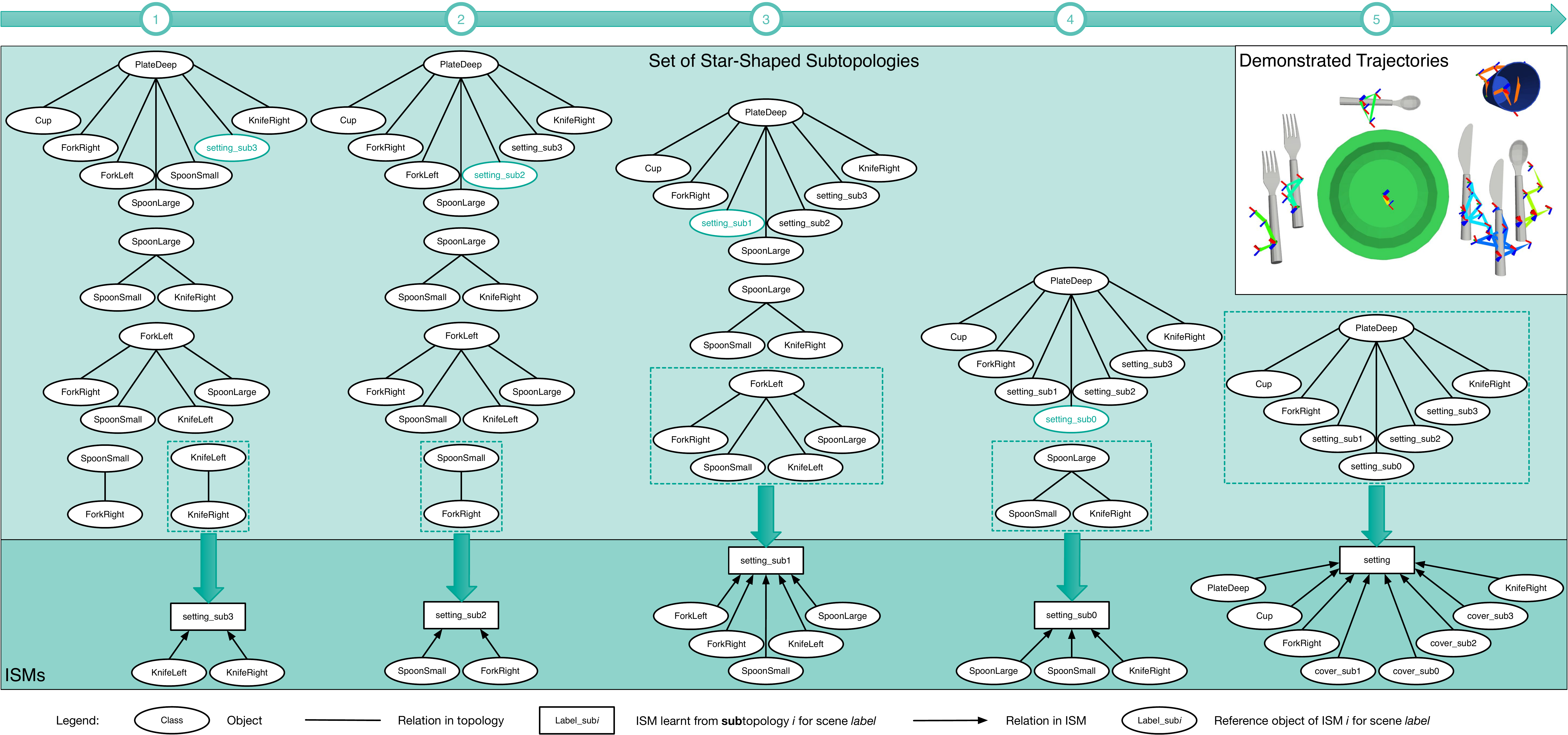}
      \caption{Algorithm - How an ISM tree is generated from the star topologies shown in the leftmost column (see 1). In each of the five depicted iterations, a star topology is selected to learn a single ISM using the object trajectories for scene category ``Setting-Ready for Breakfast''. Dashed boxes show the selected stars, whereas arrows link these stars to the learned ISMs.}
  \label{fig:psr_learn_ism_tree}
\end{figure*}

We illustrate how this deep-first search works in \figurename~\ref{fig:psr_factor_out} using the scene category ``Setting-Ready for Breakfast'' whose connected relation topology was partitioned into five star topologies in five iterations \(j \in \{1,\dots,5\}\). In video clip 1 (``Demonstration of object configurations for learning a scene classifier''), we provide footage from the demonstration we recorded for this scene category. The recorded dataset consists of object trajectories that are 112 time steps long. The star topology we extracted first on the left of \figurename~\ref{fig:psr_factor_out} had ``PlateDeep'' as its center\footnote{Each star topology extracted in one iteration is colored green.} and all other objects as its neighborhood. We selected the center for the next star topology to be extracted within this neighborhood. We stored the order in which objects \(o\) in the input topology \(\Sigma_{\nu}\) would have been chosen as centers for star topologies \(\Sigma_{\sigma}\) in a height function \(h_{\{\Sigma_{\sigma}\}}(o)\). This order would correspond to a breadth-first search. The height function is defined for each object and will be used as a balancing criterion when generating ISM trees in the next subsection, ensuring that the height of the generated tree is minimized. By favoring objects with high degrees, the depth-first search in this subsection ensures that as few star topologies as possible are extracted. All five star topologies extracted from the input topology for ``Setting-Ready for Breakfast'' can be seen in the leftmost column in \figurename~\ref{fig:psr_learn_ism_tree}. Since a depth-first search can completely search any connected graph or relation topology, and its search tree consists of the star topologies we want to extract, we can find a partitioning for any connected input topology.
  
\subsection{Contribution 1 --- Generation Algorithm for ISM Trees}\label{sec:psr_loismt}

\begin{figure*}[tpb]
  \centering
      \includegraphics[width=1\linewidth]{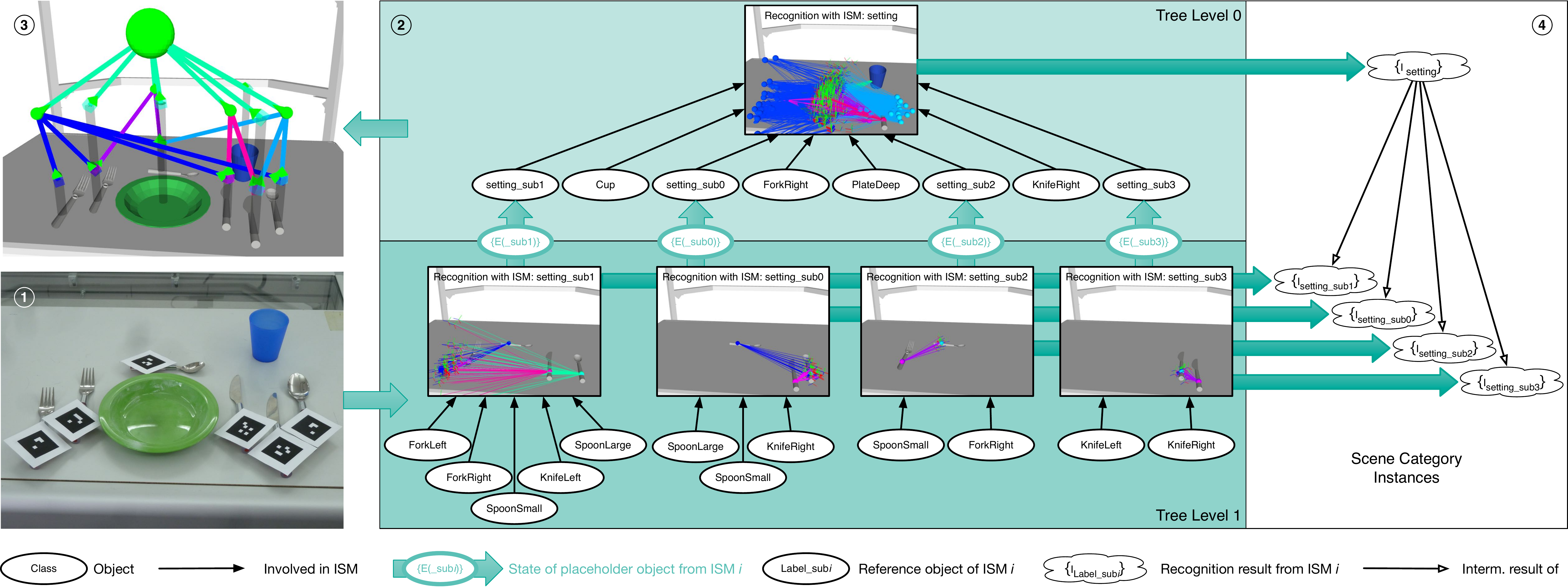}
      \caption{Algorithm - How \textbf{Passive Scene Recognition} works with an ISM tree: As soon as object poses estimated from the configuration in \textbf{1} are passed to the tree, data flows in \textbf{2} from the tree's bottom to its top, eventually yielding the scene category instance shown in \textbf{3}. This instance is made up of the recognition results in \textbf{4}, returned by the single ISMs.}
  \label{fig:psr_ism_tree_recog_interrel}
\end{figure*}

Having obtained a set of star topologies in step one, the task in step two is to generate an ISM tree from them. As one of the three contributions of this article, the algorithm we present here models all extracted star topologies by separate ISMs \(m\) which must, however, be linked together to form a tree. Such tree, generated from the five star topologies shown in the upper left in \figurename~\ref{fig:psr_learn_ism_tree}, is visualized as a directed graph in \figurename~\ref{fig:psr_ism_tree_recog_interrel}. This tree consists of a set \(\{m\}\) of five connected ISMs arranged in two levels. At the top of the tree is the root ISM \(m_{R}\), where intermediate results are merged from the four other ISMs \(m \in \{m\}\) below. All results we obtain from single ISMs in the tree are hereinafter referred to as recognition results \(\textbf{I}_{m}\). We use this term to distinguish between results of single ISMs and the instances \(\textbf{I}_{\textbf{S}}\) of a scene category \(\textbf{S}\) that result from recognizing scenes with an entire ISM tree.

Within an ISM tree, we also distinguish between real objects found at the leaves \(o_{L}\) and placeholder objects \(o_{F}\) found at the internal vertices of the tree.\footnote{Leaves and internal vertices are both represented as circles in \figurename~\ref{fig:psr_ism_tree_recog_interrel}. Internal vertices are named after the scene category and connected to ISMs by green arrows. ISMs are visualized as boxes.} In \figurename~\ref{fig:psr_ism_tree_recog_interrel}, all inputs for the ISMs at tree level 1 are leaves. Each ISM at this level models relations between real objects and a reference object \(o_{F}\), but compared to Sec. \ref{sec:psr_co_2} this reference is now an placeholder object in its own right. This placeholder object is used as an interface to pass recognition results \(\textbf{I}_{m}\) of an ISM \(m\) to another ISM \(m' \in \{m\}\) at the next lower level in the tree for further processing. In \figurename~\ref{fig:psr_ism_tree_recog_interrel}, the reference object ``setting\_sub1'' is used to pass results from the ISM on the lower left to the root ISM. In the root ISM, this reference object is treated as a regular object whose relation to another objects is modeled.

Step two generates ISM trees through two nested loops\footnote{Pseudocode is provided in the appendix by Algo. \ref{alg:psr_generateISMTreeFromSubTopologies}.}. An outer loop converts a star topology into a single ISM in each iteration step, while an inner loop is responsible for attaching the just generated ISM at the appropriate place in the tree. How this is done for our ongoing example on the scene category ``Setting-Ready for Breakfast'' can be seen in \figurename~\ref{fig:psr_learn_ism_tree}. In this figure, the iterations of the outer loop are visualized column by column from left to right, while the inner loop traverses the star topologies in each column from top to bottom. The order in which the outer loop selects star topologies from the previously extracted set is given by the height function \(h_{\{\Sigma_{\sigma}\}}(o)\) from the previous subsection. This function allows the star topologies \(\Sigma_{\sigma}(j)\) to be processed in the reverse order in which they would have been extracted in a breadth-first search. This ensures that star topologies that could be located in the highest levels of the tree to be generated are attached as close to the root as possible. This minimizes the actual height of the tree. On the left in \figurename~\ref{fig:psr_learn_ism_tree}, the last extracted star topology with ``KnifeLeft'' as its center is accordingly converted into an ISM in the first iteration of the outer loop.\footnote{The respective selected star topology is surrounded by a dashed rectangle.} The conversion is done using the ISM learning technique from Sec. \ref{sec:psr_co_2} (\cite{meissner2013}). A single ISM \(m\) is created from this topology and the trajectories \(\textbf{J}(o)\) demonstrated. Such ISM is shown in \figurename~\ref{fig:psr_learn_ism_tree} on the left of the lower dark green area.

Before the next iteration of the outer loop can begin, the inner loop still has to answer the question to which ISM \(m'\) the newly created ISM \(m\) should be connected. This connection is made utilizing the placeholder reference object \(o_{F}\) of ISM \(m\). Two ISMs \(m\) and \(m'\) can be connected only if their respective star topologies \(\Sigma_{\sigma}(j)\) and \(\Sigma_{\sigma}(k)\) have an object in common. The connection is created by replacing such a common object in the neighborhood \(N(o_{M})\) of the center \(o_{M}\) of the latter star topology \(\Sigma_{\sigma}(k)\) with the reference object \(o_{F}\) of the ISM \(m\) for the former star topology \(\Sigma_{\sigma}(j)\). To be able to later learn ISM \(m'\) with the technique from \cite{meissner2013}, the trajectory \(\textbf{J}(o)\) demonstrated for the common object is replaced with an placeholder trajectory \(\textbf{J}(o_{F})\) for the reference object \(o_{F}\) of ISM \(m\). To minimize the height of the resulting ISM tree, the inner loop starts its search for a star topology \(\Sigma_{\sigma}(k)\) suitable for this substitution, at the topologies that minimize the height function \(h_{\{\Sigma_{\sigma}\}}(o)\) and thus would be located as close to the root as possible. In the leftmost column in \figurename~\ref{fig:psr_learn_ism_tree}, the center ``KnifeLeft'' of the just selected star topology \(\Sigma_{\sigma}(j)\) is found in the topmost topology \(\Sigma_{\sigma}(k)\) which has ``PlateDeep'' as its center. In the topmost topology, ``KnifeLeft'' is replaced by the reference object ``setting\_sub3'' of the ISM just created.\footnote{Substitutions by reference objects in star topologies are colored green.}

\begin{figure*}[tpb]
  \centering
      \includegraphics[width=1\linewidth]{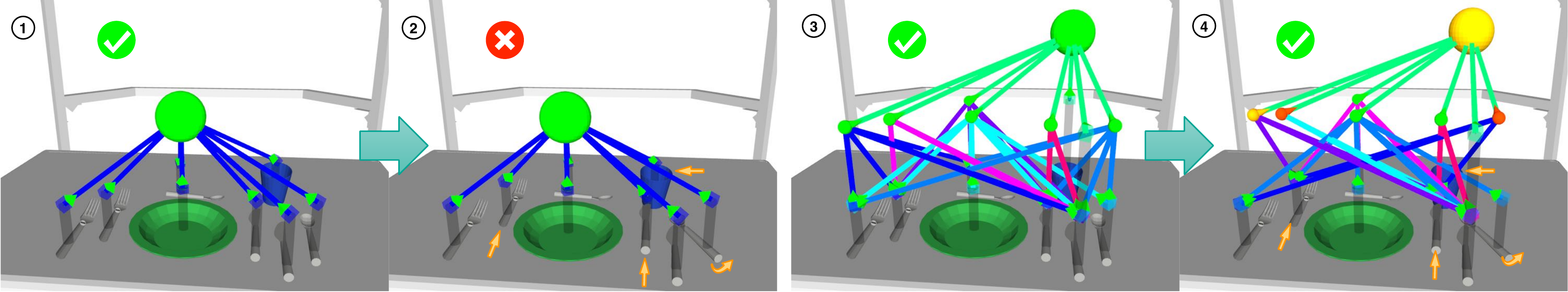}
      \caption{Results from ISM trees for the same demonstration, but learned with different topologies. Valid results are annotated with a tick in green, invalid results by a cross in red. In \textbf{1} \& \textbf{2}, a star is used. In \textbf{3} \& \textbf{4}, a complete topology. In \textbf{1}, \textbf{3}, valid object configurations are processed. Instead, \textbf{2}, \textbf{4} show invalid configurations.}
  \label{fig:psr_complete_top_for_ism_tree}
\end{figure*}

\subsection{Contribution 2 --- Recognition Algorithm for ISM Trees}\label{sec:psr_rwismt}

We concretize our definition of scene recognition from Sec. \ref{sec:psr_srd} as follows for ISM trees: From an object configuration such as the table setting in \textbf{1} in \figurename~\ref{fig:psr_ism_tree_recog_interrel}, more specifically from the estimated states\footnote{The object pose estimation is omitted in \figurename~\ref{fig:psr_ism_tree_recog_interrel} for simplicity.} \(\{\textbf{E}(o,t)\}\) of the objects, we want to derive instances \(\textbf{I}_{\textbf{S}}\) of a scene category \(\textbf{S}\), like the one shown in \textbf{3} in \figurename~\ref{fig:psr_ism_tree_recog_interrel}. Our algorithm for scene recognition using ISM trees is another contribution of this article and involves two steps: An evaluation step exemplified in \textbf{2} in \figurename~\ref{fig:psr_ism_tree_recog_interrel} and an assembly step exemplified in \textbf{4} in \figurename~\ref{fig:psr_ism_tree_recog_interrel}. Both steps\footnote{Pseudocodes for the evaluation and assembly steps are provided in the appendix by Algo. \ref{alg:psr_evaluateISMsInTree} to \ref{alg:psr_findSubInstances}.} are detailed in this subsection. In the evaluation step, all the single ISMs \(m\) in a tree are evaluated one by one, e.g., the five ISMs in the example tree in \textbf{2}, and all their respective recognition results  \(\textbf{I}_{m}\) are stored for the assembly step. In the assembly step, the recognition results from different ISMs, that belong to the same instance of a scene category are combined.

The evaluation step solves two problems: It defines an order in which the ISMs are evaluated as well as the use of an interface to exchange recognition results between ISMs. The actual evaluation of each ISM draws on the technique for classifying scenes with a single ISM. It is from our previous work (\cite{meissner2013}) and outlined in Sec. \ref{sec:psr_co_2}. In an ISM tree, the ISMs cannot all be evaluated simultaneously, since some ISMs \(m'\) are supposed to further process the intermediate results of other ISMs \(m\). These connections between pairs of ISMs, induced by reference objects \(o_{F}\), must be taken into account. For example, the evaluation of root ISM \(m_{R}\) (visualized as a box at tree level 0 in \textbf{2} in \figurename~\ref{fig:psr_ism_tree_recog_interrel}) cannot begin until the evaluation of all four ISMs \(m_{k}\) with \(k \in \{0,\dots,3\}\) at tree level 1 (the dark green area) is completed. By considering these connections, the evaluation step maximizes efficiency because each ISM is evaluated exactly once during scene recognition.

The evaluation step begins by sorting the ISMs according to their levels in the tree. This sorted list is traversed using two nested loops such that all recognition results from ISMs at tree level \(n\) can be stored before the evaluation of the ISMs at tree level \(n-1\) begins. In \textbf{2} in \figurename~\ref{fig:psr_ism_tree_recog_interrel}, this equates to evaluating the ISMs from bottom to top line by line. If only real objects, i.e., only leaves \(o_{L}\) and no internal vertices, are involved in the evaluation of the ISMs at a certain level, it is sufficient that the evaluation step distributes the different object states \(\{\textbf{E}(o,t)\}\) that describe the object configuration to the appropriate ISMs. At tree level 1 in \textbf{2}, for instance, this is the case. If internal vertices are involved in an ISM, reference objects \(o_{F}\), more precisely their placeholder states \(\textbf{E}(o_{F})\), should be computed before the ISM's evaluation. For instance, to evaluate root ISM \(m_{R}\) at level 0 in \textbf{2}, such placeholder states should be derived for the reference objects ``setting\_sub\(k\)'' from all recognition results returned by the ISMs \(m_{k}\) and passed to the root. These placeholder states are visualized as vertical green arrows emanating from the ISMs from which they originate and pointing to the internal vertex where they are further processed. Each placeholder state includes a pose \(\textbf{T}_{F}\) which is the pose of a recognition result returned by an ISM. Such pose is the location in the ISM's accumulator at which the recognition result (a highly rated combination of votes) has been identified during the evaluation of the ISM.

Each ISM \(m\) that is not the root ISM may pass zero to a multitude of recognition results to another ISM \(m'\) in the tree. When the ISM \(m'\) is evaluated, each of these results is considered as a separate input, which yields more recognition results in this ISM. These results should be passed on to a third ISM. We implemented two strategies that mitigate this effect to avoid a combinatorial explosion in scene recognition: Firstly when we generate ISM trees, the height function \(h_{\{\Sigma_{\sigma}\}}(o)\) is used to minimize tree heights and thus the lengths of the chains of interdependent ISMs in a tree. Secondly, the number of placeholder states \(\textbf{E}(o_{F})\) emanating from each ISM is limited by discarding all recognition results that have been assigned too low confidence levels \(b(o_{F})\).

The evaluation step ends once it has evaluated the ISM at the root. The recognition results from the different ISMs are visualized as clouds in \textbf{4} in \figurename~\ref{fig:psr_ism_tree_recog_interrel}. The results are connected through horizontal green arrows to those ISMs where they were computed. The task of the assembly step that now begins is to determine across ISMs which recognition results belong to the same instance \(\textbf{I}_{\textbf{S}}\) of a scene category and to assemble such instances. As in \textbf{4}, the assembly step starts at the results of the root. It recursively compares from top to bottom the stored recognition results \(\textbf{I}_{m}\), \(\textbf{I}_{m'}\) according to the connections between pairs of ISMs \(m\), \(m'\). Such recursion chain is started in \textbf{4} for each recognition result \(\textbf{I}_{\text{setting}}\) computed by root ISM \(m_{R}\). During each recursion chain, a recognition result \(\textbf{I}_{\text{setting}}\) is compared with the intermediate results \(\textbf{I}_{\text{setting\_sub}k}\) of the different ISMs \(m_{k}\) at level 1. For a comparison to assign two recognition results \(\textbf{I}_{m}\), \(\textbf{I}_{m'}\) to the same instance, two conditions must be met: Firstly these results must come from two ISMs \(m\), \(m'\) that exchanged reference objects \(o_{F}\). Secondly the very same reference object must have been involved in both recognition results. The second condition is satisfied if one of the reference objects in each of the two recognition results \(\textbf{I}_{m}\), \(\textbf{I}_{m'}\) has the same state \(\textbf{E}(o_{F})\).


\subsection{Relation Topology Selection --- Previous Work}\label{sec:rts_co}

While we explained how we partition relation topologies in Sec. \ref{sec:psr_rtp}, we did not address how to determine the relation topology to partition. Our novel generation algorithm can derive an ISM tree for any kind of connected topology, but not every topology is equally suitable for learning a classifier. \figurename~\ref{fig:psr_complete_top_for_ism_tree} illustrates how omitting the wrong relations can lead to recognition errors, i.e., false-positive results. We define a scene category instance \(\textbf{I}_{\textbf{S}}\) to be a false positive if scene recognition assigns it a confidence level \(b(\textbf{I}_{\textbf{S}})\) that exceeds a given threshold, whereas its underlying object configuration \(\{o'\}\) does not sufficiently match that scene category. \textbf{1} in \figurename~\ref{fig:psr_complete_top_for_ism_tree} visualizes a result of scene recognition for the ``Setting-Ready for Breakfast'' category. To generate the tree employed here, we used a star topology whose center is the green plate.

From a valid place setting (as in \textbf{1} and \textbf{3} in \figurename~\ref{fig:psr_complete_top_for_ism_tree}), we expect that utensils such as forks, knives and spoons be on the ``correct'' sides of the plate. In addition to this first set of rules, others require that forks, knives, and spoons be oriented parallel to each other. There are also rules regarding the relative distances of utensils from the edge of the table. If a star topology is now used to cover the first set of rules, the other rules cannot be modeled with this topology. For this reason, the ISM tree from a star topology already used in \textbf{1} in \figurename~\ref{fig:psr_complete_top_for_ism_tree} returns a false positive in \textbf{2}. The invalid configuration \(\{o'\}\) shown in \textbf{2} differs from the valid \(\{o\}\) in \textbf{1} in that several relative poses between object pairs that do not involve the plate are invalid. The ISM tree, however, does not notice these differences visualized by yellow arrows in \textbf{2}. \footnote{This false positive is indicated by a white cross on a red background, whereas white check marks on a green background indicate true positives.}

\begin{figure*}[tpb]
  \centering
      \includegraphics[width=1\linewidth]{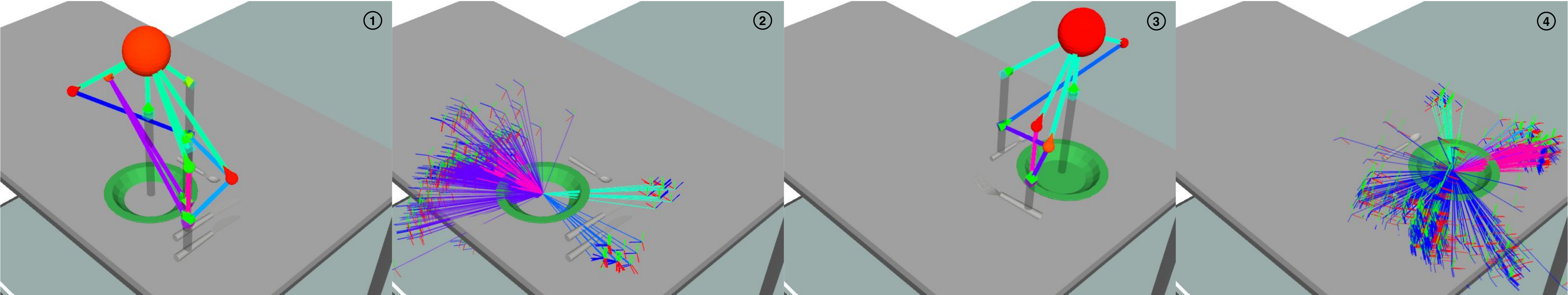}
  \caption{Example of how scene recognition (see \textbf{1}, \textbf{3}) and object pose prediction (see \textbf{2}, \textbf{4}) can compensate for changing object poses. The results they return are equivalent, even though the localized objects are rotated by \({90\,}^{\circ}\) between \textbf{1}, \textbf{2} and \textbf{3}, \textbf{4}.}
  \label{fig:asr_pose_prediction}
\end{figure*}

To prevent false positives, ISM trees could instead be learned from all \(\sfrac{n \cdot (n-1)}{2}\) spatial relations that can be defined for a set of \(n\) objects, i.e., from a complete relation topology. The fact that ISM trees from such complete topologies do not yield false positives is illustrated in \textbf{3} and \textbf{4} in \figurename~\ref{fig:psr_complete_top_for_ism_tree}. In \textbf{3} and \textbf{4}, such a tree is applied to the object configurations \(\{o\}\), \(\{o'\}\) from \textbf{1}, \textbf{2} in \figurename~\ref{fig:psr_complete_top_for_ism_tree}. The result in \textbf{4} is not a false positive, as some of the ISMs in the tree recognize that some of the relations modeled by them are not fulfilled. \footnote{The colors of the spheres above the single ISMs in the tree indicate to which degree their respective relations are fulfilled.} However, a disadvantage of complete topologies is the excessive number of relations that must be checked during scene recognition. In general, the cost of scene recognition with ISM trees is closely related to the number of relations represented. The fact that recognition with complete topologies is generally intractable has also been reported (\cite{grauman2011visual}) for other part-based models. 

The question arose how to find a connected topology, different from the edge cases which are the star and complete topologies, as a middle ground. Such topology would yield an ISM tree that combines efficiency and representational power. To identify such a relation topology most generically, we used two domain-unspecific goodness measures in our previous work (\cite{meissner2015}): The false-positive rate numFPs$()$ in scene recognition and the average time consumption avgDur$()$ of scene recognition. Based on these measures, we formalized the selection of relation topologies as a combinatorial optimization problem. The challenge in this selection is the exponential number \(2^{\sfrac{n \cdot (n-1)}{2}}\) of relation topologies that can be defined for \(n\) objects. Given the number of topologies among which to choose, we used a local search technique to develop a Relation Topology Selection procedure. Its basic idea was to iteratively adjust a relation topology by adding, removing, or exchanging relations until a topology was found that contained only those relations that were most important for recognizing a scene category. The result, a so-called optimized topology, was then used to learn an ISM tree from it.

\section{Methods --- Active Scene Recognition}\label{sec:asr}

\subsection{State Machine and Next-Best-View --- Previous Work}\label{sec:asr_co}

In the previous section on Passive Scene Recognition (PSR), we ignored the question of under which conditions object pose estimation can obtain ``object estimates'' for scene recognition. Our approach to creating suitable conditions in spatially distributed and cluttered indoor environments is to have a mobile robot adopt camera views from which it can perceive searched objects. To this end, in two previous works (\cite{meissner2014}, \cite{meissner2016}), we introduced Active Scene Recognition (ASR) --- an approach that connects PSR with three-dimensional object search within a decision-making system. We implemented ASR as a state machine consisting of two search modes (states) DIRECT\_SEARCH and INDIRECT\_SEARCH that alternate. We then integrated this state machine with the MILD robot shown in \figurename~\ref{fig:int_intro} so that ASR can decide on the presence of \(n\) scene categories in the environment visible in \figurename~\ref{fig:int_pano}.

\begin{figure*}[tpb]
  \centering
      \includegraphics[width=1\linewidth]{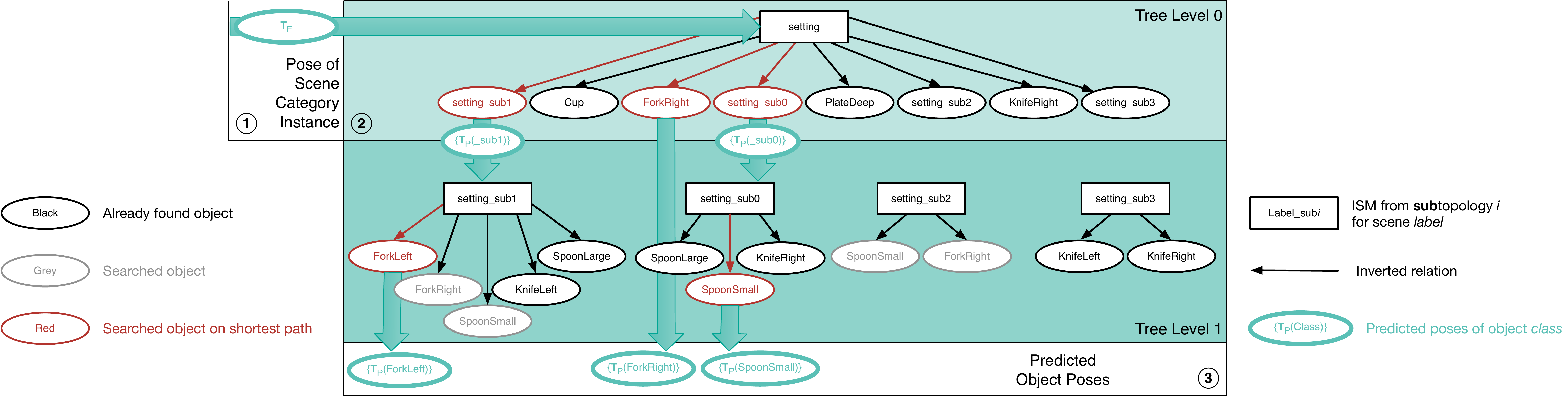}
  \caption{Algorithm - How the poses of searched objects are predicted with an ISM tree: The incomplete scene category instance in \textbf{1} is input and causes data to flow through the tree in \textbf{2}. Unlike scene recognition, data flows from the tree's top to its bottom, while relative poses from different inverted spatial relations are combined. The resulting predictions are visible in \textbf{3}.}
  \label{fig:asr_data_flow_during_prediction}
\end{figure*}

ASR starts in DIRECT\_SEARCH mode which is tasked with acquiring initial object estimates. For this purpose, we developed two strategies to identify suitable camera views and move to them. The first (``informed'') strategy is based on prior knowledge about possible placements of objects, e.g, from demonstrations of scene categories. If this informed search does not yield object estimates, an uninformed strategy (\cite{ye1999sensor}) is used to explore the entire environment uniformly. As soon as at least one object estimate is obtained, the direct search stops, and the INDIRECT\_SEARCH mode starts instead.

The other mode INDIRECT\_SEARCH consists of a loop in which three substates (Passive Scene Recognition, a technique for predicting the poses of searched objects, and 3-D object search) alternate. The loop starts in the first substate SCENE\_RECOGNITION, in which PSR is performed with ISM trees on the currently available object estimates. The results of SCENE\_RECOGNITION are instances of scene categories. Some instances may not contain all objects belonging to their category. Therefore, it is the task of the other two substates in the loop to complete such partial instances. The second substate OBJECT\_POSE\_PREDICTION uses ISM trees to predict locations of objects that would allow completion of these instances. When using ISM trees, some object poses may need to be predicted using entire sequences of spatial relations. This is prone to a combinatorial explosion: An algorithm presented in our previous work (\cite{meissner2014}) for predicting poses suffered from such an explosion. In the next subsection, we address this problem with an efficient algorithmic solution (one of the contributions of this article).

The third substate RELATION\_BASED\_SEARCH of the loop uses predicted object poses to search for these objects in 3-D, i.e., to determine camera views that are promising for finding them. Whenever such a view has been determined, the robot moves there and tries to localize objects in 6-DoF. In \cite{meissner2016}, we formalised finding suitable camera views as a Next-Best-View (NBV) optimisation problem. The algorithm with which we addressed this problem had to search for a camera view that maximized an objective function, starting from predicted object poses and the current robot pose. This objective function modeled the success probability of object localization as well as the time required to reach the view and perform localization. Our approach allowed both optimizing the views and deciding which objects to search in them.

\subsection{Contribution 3 --- Object Pose Prediction Algorithm}\label{sec:asr_popomo}


Our approach to predicting the poses of searched objects with the help of ISM trees is similar to an inversion of scene recognition. Scene recognition infers from known states of objects which instances of a scene category \(\textbf{S}\) the states correspond to. Instead, object pose prediction infers hypotheses about the possible poses \(\textbf{T}_{P}\) of the missing objects \(o_{P}\) from a known scene category instance \(\textbf{I}_{\textbf{S}}\) and its location \(\textbf{T}_{F}\). Since these predicted poses must be suitable for 3-D object search, both the 3-DoF positions of the missing objects and their 3-DoF orientations must be predicted. Knowledge about the expected orientation of a searched object can determine the success or failure of object localization. The poses predicted by the algorithm presented in this subsection are visualized as coordinate systems in \textbf{2} and \textbf{4} in \figurename~\ref{fig:asr_pose_prediction}. ISM trees allow us to infer object poses from spatial relations \(\textbf{R}\), i.e., depending on the known poses \(\textbf{T}\) of already found objects \(o\). The flexibility of this approach is illustrated in \figurename~\ref{fig:asr_pose_prediction} for the scene category ``Setting-Ready for Breakfast'': If an incomplete instance of this scene category is rotated as between \textbf{1} and \textbf{3}, the object poses predicted from them in \textbf{2} and \textbf{4} rotate with it without need for adjustments.

\begin{figure*}[tpb]
  \centering
      \includegraphics[width=\linewidth]{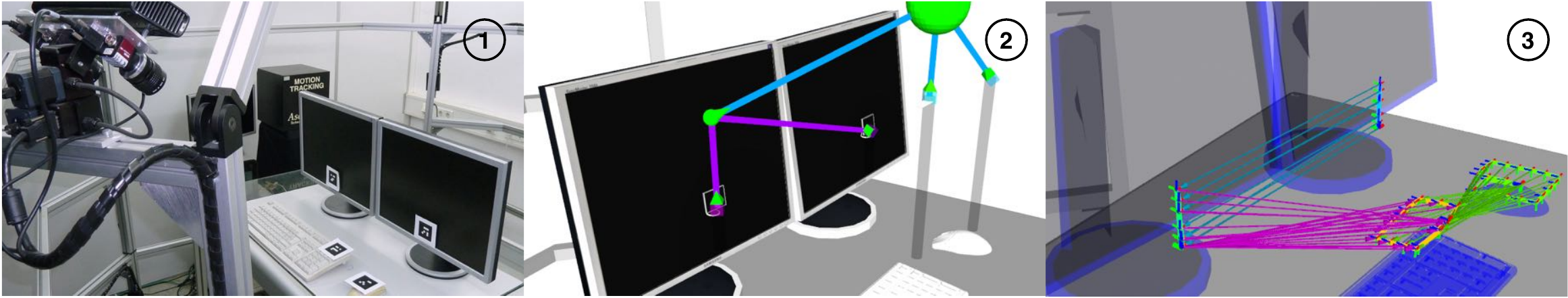}
      \caption{\textbf{1}: Snapshot of the demonstration for scene category ``Office''. \textbf{3}: The ISM tree for this category, including the relative poses in its relations and the demonstrated poses. \textbf{2}: Result of applying this tree onto the configuration ``Correct-configuration''.}
  \label{fig:eva_office_2016_model}
\end{figure*}

Predicting object poses with ISM trees consists of two steps and is the third and final contribution of this article: In step one, precomputations are performed to identify those parts of an ISM tree that provide a fast and reliable prediction. In step two, these precomputations are used to predict the poses of searched objects. \figurename~\ref{fig:asr_data_flow_during_prediction} refers to the ISM tree which models scene category ``setting'' and has already been used in \figurename~\ref{fig:psr_ism_tree_recog_interrel}. Since some objects from the scene category are involved in multiple relations, several leaves \(o_{L}\) in the tree correspond to the same object. This way, ``ForkRight'' is represented at both levels of the tree. To predict object poses using the leaf for ``ForkRight'' at tree level 1, one would have to combine spatial relations from the ISMs ``setting'' and ``setting\_sub1''. On level 0, a single relation in the ISM ``setting'' is sufficient. Since the accuracy of predicted poses depends on the number of relations used, step one precomputes the shortest sequences of ISMs between any object \(o\) in a scene category \(\textbf{S}\) and the root \(m_{R}\) of the tree. All nontrivial sequences are defined as paths \(\textbf{P}(m_{R}, o)\) consisting of \(l\) pairs \((m_{k},m_{k+1})\) of connected ISMs (see Eq. \ref{eq:asr_Eq1}). Paths end at the ISM \(m_{l+1}\) which contains the leaf \(o_{L}\) appropriate for predicting the pose of object \(o\). We compute the shortest paths using a breadth-first search that traverses trees, as in \figurename~\ref{fig:asr_data_flow_during_prediction}, from top to bottom.

\begin{equation}\label{eq:asr_Eq1}
  \footnotesize
  \begin{split}
    \textbf{P}(m_{R}, o) = \left\{ (m_{k},m_{k+1}) \left| \right. \right. & m_{1} = m_{R} \wedge \forall k: m_{k} \text{ connected with}\\
& m_{k+1} \text{ by an } o_{F} \wedge \left. o \text{ is a }  o_{L} \text{ of } m_{l+1} \right\}
    \end{split}
\end{equation}

Step two, the actual pose prediction algorithm, derives possible poses for the searched objects from these paths and a partial scene category instance \(\textbf{I}_{\textbf{S}}\). This is done via three nested loops: The innermost loop\footnote{Pseudocode for the innermost loop is provided in the appendix by Algo. \ref{alg:asr_predictPose}.} computes exactly one pose estimate \(\textbf{T}_{P}\) per searched object \(o_P\). Two outer loops\footnote{Pseudocode for the outer loops is provided in the appendix by Algo. \ref{alg:asr_generateCloudOfPosePredictions}.} call this innermost loop until a specified number \(n_{P}\) of poses is predicted for each searched object. \figurename~\ref{fig:asr_data_flow_during_prediction} shows how the innermost loop operates on an ISM tree. First, as shown by the horizontal green arrow in \textbf{1}, it passes the pose \(\textbf{T}_{F}\) of instance \(\textbf{I}_{\textbf{S}}\) to the root ISM. Starting from root ISM \(m_{R}\) in \textbf{2}, it evaluates all ISMs on the shortest path \(\textbf{P}(m_{R}, o_P)\) to a suitable leaf. For ``ForkLeft'', such a leaf is located on the far left of tree level 1. A predicted pose is visualized in \textbf{3} as a green circle and connected by a green arrow to the leaf from which it results.

The algorithm originating from our previous work (\cite{meissner2014}) was unable to efficiently predict object poses because it processed all the relative poses that make up a spatial relation in an ISM. Across multiple ISMs, this would lead to a combinatorial explosion: If it generated a prediction for each relative pose in a relation of an ISM \(m_{k}\) at tree level \(k\) and passed the prediction as a possible pose of a reference object to another ISM \(m_{k+1}\) at level \(k+1\), each of these would be combined with all relative poses in a relation of ISM \(m_{k+1}\). Also, the algorithm did not use shortest paths. Combinatorial explosion is avoided in the new method presented here by processing one random relative pose per relation instead of all. More precisely, the innermost loop selects one relative pose \(\textbf{T}_{jk}\) from each ISM along the shortest path and inverts all such poses. \footnote{To visualize in \figurename~\ref{fig:asr_data_flow_during_prediction} that spatial relations have been inverted to predict object poses, all arrows point from top to bottom instead of bottom to top. The shortest paths used to predict poses are colored red.} The pose \(\textbf{T}_{F}\) of the incomplete instance is multiplied by all these sampled and inverted relative poses \(\textbf{T}_{kj}\) so that one of the sought pose hypotheses \(\textbf{T}_{P}\) is obtained. For instance, to predict an absolute pose of ``ForkLeft'', the innermost loop randomly selects one relative pose from the ISMs ``setting'' and ``setting\_sub1'' respectively.

\begin{table*}[tpb]
\scriptsize
\centering
\caption{We specify position and orientation compliances, as well as a similarity rating for all measured object poses, which ISM ``Office\_sub0'' processes. The value of the recognition result's objective function (returned by the ISM) is also provided. See Sec. \ref{sec:ear_eopsr_pdsci} for definitions of the objective function, rating, and compliances.}
\begin{tabular}{@{}llllllllll@{}}
\toprule
Object Configuration & \multicolumn{3}{c}{LeftScreen} & \phantom{a} & \multicolumn{3}{c}{RightScreen} & Obj. Function \\ \cmidrule{2-4}\cmidrule{6-8}
 & Simil. & Pos. & Orient. && Simil. & Pos. & Orient. &  \\
\midrule
Correct-configuration          & \textcolor{cdarkgreen}{1.00} & \textcolor{cdarkgreen}{1.00} & \textcolor{cdarkgreen}{1.00} && \textcolor{cdarkgreen}{1.00} & \textcolor{cdarkgreen}{1.00} & \textcolor{cdarkgreen}{1.00} & \textcolor{cdarkgreen}{2.00} \\
RightScreen-half-lowered  & \textcolor{cdarkgreen}{1.00} & \textcolor{cdarkgreen}{1.00} & \textcolor{cdarkgreen}{1.00} && \textcolor{cdarkgreen}{0.72} & \textcolor{cdarkgreen}{0.72} & \textcolor{cdarkgreen}{1.00} & \textcolor{cdarkgreen}{1.72} \\
RightScreen-fully-lowered & \textcolor{cdarkyellow}{0.00} & \textcolor{cdarkyellow}{0.00} & \textcolor{cdarkyellow}{0.00} && \textcolor{cdarkyellow}{1.00} & \textcolor{cdarkyellow}{1.00} & \textcolor{cdarkgreen}{1.00} & \textcolor{cdarkgreen}{1.00} \\
LeftScreen-half-front        & \textcolor{cdarkyellow}{1.00} & \textcolor{cdarkyellow}{1.00} & \textcolor{cdarkgreen}{1.00} && \textcolor{cdarkyellow}{0.60} & \textcolor{cdarkyellow}{0.61} & \textcolor{cdarkgreen}{0.98} & \textcolor{cdarkgreen}{1.60} \\
RightScreen-half-right      & \textcolor{cdarkgreen}{1.00} & \textcolor{cdarkgreen}{1.00} & \textcolor{cdarkgreen}{1.00} && \textcolor{cdarkgreen}{0.96} & \textcolor{cdarkgreen}{0.97} & \textcolor{cdarkgreen}{0.99} & \textcolor{cdarkgreen}{1.96} \\
LeftScreen-half-rotated     & \textcolor{cdarkgreen}{0.54} & \textcolor{cdarkgreen}{0.93} & \textcolor{cdarkgreen}{0.58} && \textcolor{cdarkgreen}{1.00} & \textcolor{cdarkgreen}{1.00} & \textcolor{cdarkgreen}{1.00} & \textcolor{cdarkgreen}{1.54} \\
LeftScreen-fully-rotated    & \textcolor{cdarkgreen}{0.00} & \textcolor{cdarkgreen}{0.00} & \textcolor{cdarkgreen}{0.00} && \textcolor{cdarkgreen}{1.00} & \textcolor{cdarkgreen}{1.00} & \textcolor{cdarkgreen}{1.00} & \textcolor{cdarkgreen}{1.00} \\
Mouse-half-right              & \textcolor{cdarkgreen}{1.00} & \textcolor{cdarkgreen}{1.00} & \textcolor{cdarkgreen}{1.00} && \textcolor{cdarkgreen}{0.99} & \textcolor{cdarkgreen}{1.00} & \textcolor{cdarkgreen}{0.99} & \textcolor{cdarkgreen}{1.99} \\
Mouse-half-rotated          & \textcolor{cdarkgreen}{1.00} & \textcolor{cdarkgreen}{1.00} & \textcolor{cdarkgreen}{1.00} && \textcolor{cdarkgreen}{0.99} & \textcolor{cdarkgreen}{0.99} & \textcolor{cdarkgreen}{1.00} & \textcolor{cdarkgreen}{1.99} \\
\bottomrule
\end{tabular}
\label{tab:eva_office2016_recognition_values_sub0}
\end{table*}

\begin{table*}[tpb]
\scriptsize
\centering
\caption{We specify position and orientation compliances, as well as a similarity rating for all measured object poses, which ISM ``Office'' processes. The value of the recognition result's objective function (returned by the ISM) is also provided. See Sec. \ref{sec:ear_eopsr_pdsci} for definitions of the objective function, rating, and compliances.}
\begin{tabular}{@{}lllllllllllll@{}}
\toprule
Object Configuration & \multicolumn{3}{c}{Keyboard} &\phantom{a} & \multicolumn{3}{c}{Mouse} &\phantom{a} & \multicolumn{3}{c}{Office\_sub0} & Obj. Function \\ \cmidrule{2-4} \cmidrule{6-8} \cmidrule{10-12}
 & Simil. & Pos. & Orient. && Simil. & Pos. & Orient. && Simil. & Pos. & Orient. &  \\
\midrule
Correct-configuration          & \textcolor{cdarkgreen}{0.98} & \textcolor{cdarkgreen}{0.98} & \textcolor{cdarkgreen}{1.00} && \textcolor{cdarkgreen}{1.00} & \textcolor{cdarkgreen}{1.00} & \textcolor{cdarkgreen}{1.00} && \textcolor{cdarkgreen}{2.00} & \textcolor{cdarkgreen}{1.00} & \textcolor{cdarkgreen}{1.00} & \textcolor{cdarkgreen}{3.98} \\
RightScreen-half-lowered & \textcolor{cdarkgreen}{0.98} & \textcolor{cdarkgreen}{0.98} & \textcolor{cdarkgreen}{1.00} && \textcolor{cdarkgreen}{1.00} & \textcolor{cdarkgreen}{1.00} & \textcolor{cdarkgreen}{1.00} && \textcolor{cdarkgreen}{1.72} & \textcolor{cdarkgreen}{1.00} & \textcolor{cdarkgreen}{1.00} & \textcolor{cdarkgreen}{3.70} \\
RightScreen-fully-lowered & \textcolor{cdarkgreen}{1.00} & \textcolor{cdarkgreen}{1.00} & \textcolor{cdarkgreen}{1.00} && \textcolor{cdarkgreen}{1.00} & \textcolor{cdarkgreen}{1.00} & \textcolor{cdarkgreen}{1.00} && \textcolor{cdarkgreen}{1.00} & \textcolor{cdarkgreen}{1.00} & \textcolor{cdarkgreen}{1.00} & \textcolor{cdarkgreen}{3.00} \\
LeftScreen-half-front        & \textcolor{cdarkgreen}{0.97} & \textcolor{cdarkgreen}{0.98} & \textcolor{cdarkgreen}{0.99} && \textcolor{cdarkgreen}{1.00} & \textcolor{cdarkgreen}{1.00} & \textcolor{cdarkgreen}{1.00} && \textcolor{cdarkgreen}{1.47} & \textcolor{cdarkgreen}{0.89} & \textcolor{cdarkgreen}{0.98} & \textcolor{cdarkgreen}{3.43} \\
RightScreen-half-right      & \textcolor{cdarkgreen}{0.99} & \textcolor{cdarkgreen}{0.99} & \textcolor{cdarkgreen}{1.00} && \textcolor{cdarkgreen}{0.99} & \textcolor{cdarkgreen}{0.99} & \textcolor{cdarkgreen}{1.00} && \textcolor{cdarkgreen}{1.96} & \textcolor{cdarkgreen}{1.00} & \textcolor{cdarkgreen}{1.00} & \textcolor{cdarkgreen}{3.94} \\
LeftScreen-half-rotated    & \textcolor{cdarkgreen}{0.98} & \textcolor{cdarkgreen}{0.98} & \textcolor{cdarkgreen}{1.00} && \textcolor{cdarkgreen}{1.00} & \textcolor{cdarkgreen}{1.00} & \textcolor{cdarkgreen}{1.00} && \textcolor{cdarkgreen}{1.46} & \textcolor{cdarkgreen}{0.93} & \textcolor{cdarkgreen}{1.00} & \textcolor{cdarkgreen}{3.45} \\
LeftScreen-fully-rotated   & \textcolor{cdarkgreen}{0.93} & \textcolor{cdarkgreen}{0.95} & \textcolor{cdarkgreen}{0.97} && \textcolor{cdarkgreen}{0.95} & \textcolor{cdarkgreen}{0.98} & \textcolor{cdarkgreen}{0.96} && \textcolor{cdarkgreen}{1.00} & \textcolor{cdarkgreen}{1.00} & \textcolor{cdarkgreen}{1.00} & \textcolor{cdarkgreen}{2.87} \\
Mouse-half-right             & \textcolor{cdarkgreen}{0.99} & \textcolor{cdarkgreen}{0.99} & \textcolor{cdarkgreen}{1.00} && \textcolor{cdarkgreen}{0.86} & \textcolor{cdarkgreen}{0.86} & \textcolor{cdarkgreen}{1.00} && \textcolor{cdarkgreen}{1.99} & \textcolor{cdarkgreen}{1.00} & \textcolor{cdarkgreen}{1.00} & \textcolor{cdarkgreen}{3.83} \\
Mouse-half-rotated         & \textcolor{cdarkgreen}{1.00} & \textcolor{cdarkgreen}{1.00} & \textcolor{cdarkgreen}{1.00} && \textcolor{cdarkgreen}{0.77} & \textcolor{cdarkgreen}{1.00} & \textcolor{cdarkgreen}{0.78} && \textcolor{cdarkgreen}{1.96} & \textcolor{cdarkgreen}{0.98} & \textcolor{cdarkgreen}{0.99} & \textcolor{cdarkgreen}{3.73} \\
\bottomrule
\end{tabular}
\label{tab:eva_office2016_recognition_values_office2016}
\end{table*}

\section{Experiments and Results}\label{sec:ear}

\subsection{Overview}\label{sec:ear_o}

We do present experiments for PSR with ISM trees in Sec. \ref{sec:ear_eopsr} and for our approach to ASR in Sec. \ref{sec:ear_eoasr}. Except for explicitly labeled experiments in Sec. \ref{sec:ear_eorts_pooit}, \ref{sec:ear_eoasr_eocotata} and \ref{sec:ear_eoasr_r}, our PSR and ASR approaches are evaluated exclusively on measurements from physical sensors. The input for these real-world experiments was acquired by the pivoting sensor head of our MILD mobile robot. Our approach to ASR controlled both the sensors and actuators of this physical robot shown in \figurename~\ref{fig:int_intro}. The robot operated in our experimental setup which mimicked some aspects of a kitchen (see \figurename~\ref{fig:int_pano}). Our approaches to PSR and ASR were run on a PC with an Intel Xeon E5-1650 v3 3.50 GHz CPU and 32GB DDR4 RAM. In Sec. \ref{sec:eva_evaasr_roswodr} to \ref{sec:ear_eoasr_rosiacoc}, ISM trees for ten scene categories are used to evaluate our ASR approach. In each subsection, ASR is expected to provide estimates for all existing scenes. In addition to these ASR experiments on a physical robot, we performed an experiment in simulation in Sec. \ref{sec:ear_eoasr_eocotata} to compare the time consumption of our approach to ASR with two alternatives.

\subsection{Evaluation of Passive Scene Recognition}\label{sec:ear_eopsr}

\subsubsection{Scene Category ``Office''}\label{sec:ear_eopsr_sco}

In this experiment, we evaluate how well our scene recognition approach captures the properties of spatial relations throughout an ISM tree. We do this by investigating how changes in individual object poses affect the recognized scene category instances. The scene category used is named ``Office'' and consists of four objects with fiducial markers attached to them to maximize object localization accuracy: Mouse, Keyboard, LeftScreen, and RightScreen. Video clip 2 (``Demonstration of the scene category: Office'') shows how we demonstrated this category. \textbf{1} in \figurename~\ref{fig:eva_office_2016_model} visualizes one of the 51 object configurations included in the dataset for this demonstration. The relative poses that make up the spatial relations of the learned ISM tree are visualized in \textbf{3} in \figurename~\ref{fig:eva_office_2016_model}. The demonstration includes two relative movements between object pairs. The first relative movement involves both screens. It shall create a relation that consists of nearly identical relative poses, as depicted in the middle of \textbf{3}. The second relative movement between Mouse and Keyboard shall create a much more variable spatial relation. The tree learned for ``Office'' consists of one ISM labeled ``Office'' and another labeled ``Office\_sub0''.

\subsubsection{Parameters Describing Results of Scene Recognition}\label{sec:ear_eopsr_pdsci}

Scene recognition is performed on nine object configurations to analyze what impact changing the pose of an object has on the parameters in scene category instances. Since scene recognition is deterministic, each configuration is processed only once. In each configuration, an object pose either differs in its position or its orientation from those expected by the spatial relations in the ISM tree. \tablename~\ref{tab:eva_office2016_recognition_values_sub0} and \ref{tab:eva_office2016_recognition_values_office2016} show how the ISMs ``Office\_sub0'' and ``Office''  quantify these differences. In both tables, color coding indicates the appropriateness of the values they contain. Green stands for results we consider excellent, yellow for good results, and red indicates problems.

Each estimated scene category instance in \figurename~\ref{fig:eva_office2016_recognition} can also be represented as a set of parameters whose values can be found in the same row of the two tables. For each object, two compliance parameters express the degree to which its estimated position and orientation comply with a spatial relation in which the object is involved. Formal definitions of these compliances can be found in \cite{meissner2018}. Compliances are normalized to \([0,1]\), where 1 expresses a perfect match, and 0 represents a lower bound below which objects are excluded from scene category instances. A similarity measure is derived for each object by multiplying both compliances. Adding up all similarity measures in a table row yields the value of the objective function for an ISM \(m\), the nonnormalized equivalent to its confidence level. This value describes the extent to which all of these objects, either directly involved with the ISM \(m\) or involved with another ISM \(m'\) to which \(m\) is related, contribute to the recognition result produced by ISM \(m\).

\subsubsection{Influence of Object Poses on Passive Scene Recognition}\label{sec:ear_eopsr_ioop}

\begin{figure*}[tpb]
  \centering
      \includegraphics[width=\linewidth]{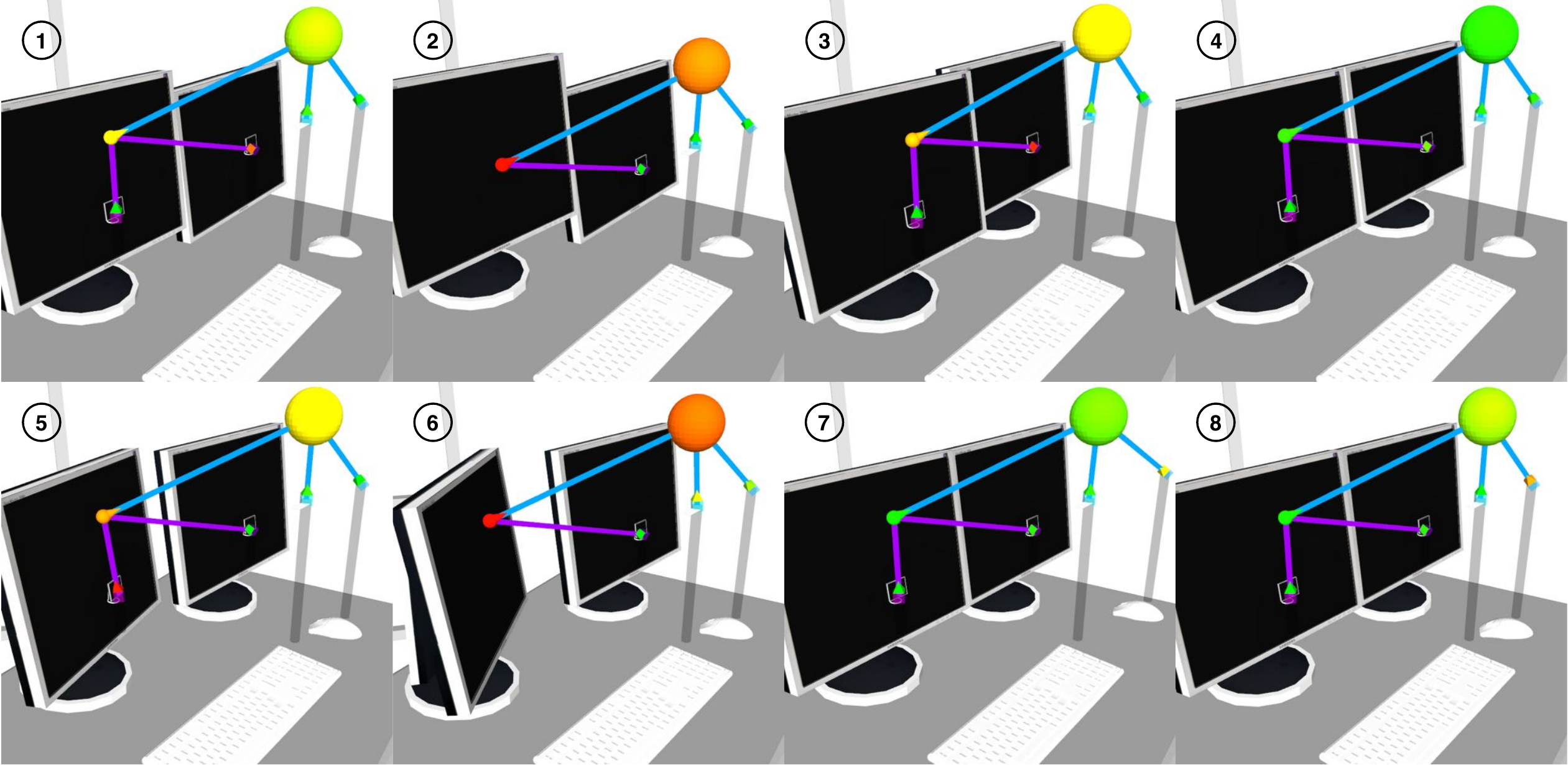}
      \caption{Visualization of the scene category instances that an ISM tree for the category ``Office'' recognized in the physical object configurations ``RightScreen-half-lowered'' (\textbf{1}), ``RightScreen-fully-lowered'' (\textbf{2}), ``LeftScreen-half-front'' (\textbf{3}), ``RightScreen-half-right'' (\textbf{4}), ``LeftScreen-half-rotated'' (\textbf{5}), ``LeftScreen-fully-rotated'' (\textbf{6}), ``Mouse-half-right'' (\textbf{7}), and ``Mouse-half-rotated'' (\textbf{8}).}
  \label{fig:eva_office2016_recognition}
\end{figure*}

In the uppermost lines of \tablename~\ref{tab:eva_office2016_recognition_values_sub0} and \ref{tab:eva_office2016_recognition_values_office2016}, all compliances concerning positions and orientations are close to one. Thus, the objective function reaches its maximum. The values of the objective function correspond to the number of objects that each ISM considers. In ``RightScreen-half-lowered'', RightScreen is moved downwards by 0.05 m, as can be seen in \textbf{1} in \figurename~\ref{fig:eva_office2016_recognition}. The compliances for ``RightScreen-half-lowered'' in \tablename~\ref{tab:eva_office2016_recognition_values_sub0} validate that ISM ``Office\_sub0'' correctly notices that RightScreen has been slided, but not rotated. To increase the discrepancy between the positions of LeftScreen and RightScreen, LeftScreen is displaced further upwards by 0.035 m in ``RightScreen-fully-lowered''. Whereas the nonzero compliances in \tablename~\ref{tab:eva_office2016_recognition_values_sub0} indicate that no object in ``RightScreen-half-lowered'' has been excluded from the scene category instance as we intended, this is different in ``RightScreen-fully-lowered''. The positional difference between the screens is sufficient to exclude one screen. However, the yellow coloring of the corresponding compliances in \tablename~\ref{tab:eva_office2016_recognition_values_sub0} makes it clear that it is suboptimal that ISM ``Office\_sub0'' excludes the less-displaced LeftScreen. ISM trees are more sensitive to displacements of reference objects of their ISMs than to those of nonreference objects.

When shifting LeftScreen forwards by 0.05 m in ``LeftScreen-half-front'' instead of moving RightScreen, scene recognition reveals that displacements are always considered from the perspective of the reference object. The two aforementioned phenomena, although counterintuitive, do not affect the values calculated for the overall objective function and could be systematically compensated. In configuration ``RightScreen-half-right'', we displace RightScreen this time. We move this screen to the right so that our experiments cover all directions in 3-D space where shifting is possible. When being moved to the right by 0.015 m, RightScreen is displaced less than in ``RightScreen-half-lowered''. The objective-function values in \tablename~\ref{tab:eva_office2016_recognition_values_sub0} show that ISMs can be sensitive enough to notice such slight differences.

After checking whether ISM trees detect translations of objects, the same should be done for rotations. We rotate LeftScreen by \({15\,}^{\circ}\) in ``LeftScreen-half-rotated'' and by \({30\,}^{\circ}\) in ``LeftScreen-fully-rotated''. Comparing the configurations in which LeftScreen is rotated with those in which RightScreen is lowered reveals that rotations are just as precisely detected as translations. After having analyzed how changing object poses affects ISM ``Office\_sub0'', the next configurations are to show that changes are treated equally in root ISM ``Office''. In the configuration ``Mouse-half-right'', Mouse is pushed 0.11 m to the right. This represents a displacement larger than those of both screens together in ``RightScreen-fully-lowered''. However, Mouse is not excluded from the corresponding scene category instance in \textbf{7} in \figurename~\ref{fig:eva_office2016_recognition}. This shows that we can control how permissive spatial relations are through the demonstrations we record. In ``Mouse-half-rotated'', Mouse is rotated by \({15\,}^{\circ}\) instead of being shifted. The fact that the objective function of ISM ``Office'' returns the same value for ``Mouse-half-rotated'' and ``LeftScreen-half-rotated'' proves that we can further influence whether scene recognition is permissive concerning positions or orientations. Overall, these experiments confirm that ISM trees can identify whether an object has been shifted or rotated in various directions. They can also estimate the sizes of such displacements. 

\subsubsection{Scene Categories Demonstrated for ASR Evaluation}\label{sec:ear_eorts_scfae}

\begin{figure*}[tpb]
  \centering
      \includegraphics[width=1\linewidth]{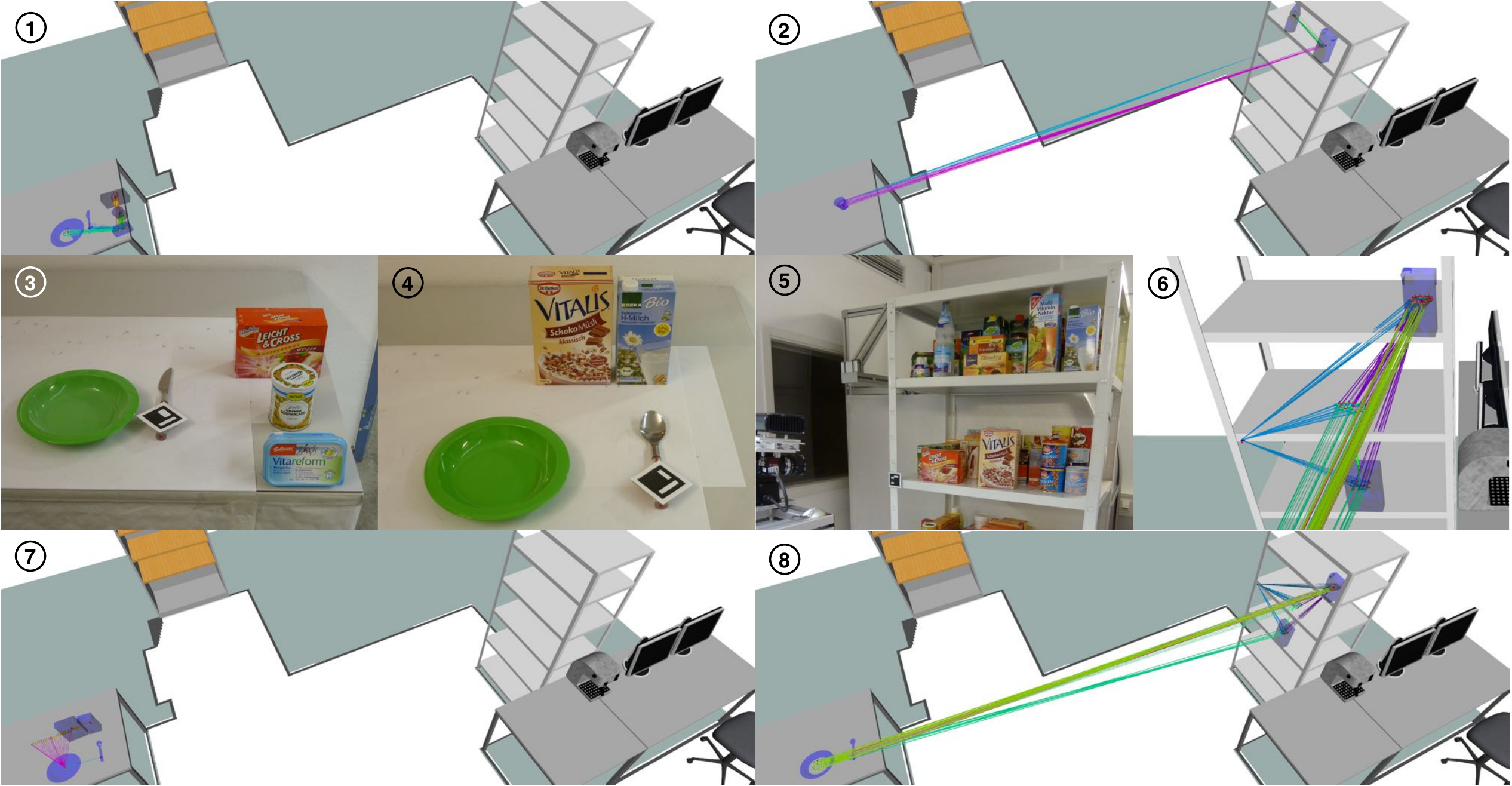}
  \caption{Object trajectories demonstrated for the scene categories ``Sandwich-Setting, ``Drinks-on Shelf'', ``Cereals-Setting'', and ``Cereals-on Shelf'' in our kitchen setup, as well as the spatial relations inside the ISM trees learned for these categories.}
  \label{fig:eva_scene_categories}
\end{figure*}

This subsection is devoted to the scene categories that we demonstrated to evaluate ASR and that are named in \tablename~\ref{tab:eva_ism_tree_performances}. The next subsection is then devoted to the performance of the ISM trees learned from the demonstrations we recorded for these categories. Demonstrations and the evaluation of ASR took place in the kitchen setup depicted in \figurename~\ref{fig:int_pano} or \textbf{1} in \figurename~\ref{fig:eva_scene_categories}. Object configurations were demonstrated in areas of the setup such as the cupboard at the top of \textbf{1}, the shelves on its right, and the tables. The cupboard and shelves are filled with clutter. We recorded all object poses with the camera head of our MILD robot. Markers are only used on the cutlery to compensate for reflections. The ISM trees for all scene categories in \tablename~\ref{tab:eva_ism_tree_performances} result from topologies optimized using the Relation Topology Selection (RTS), which we outlined in Sec. \ref{sec:rts_co}. This table contains the durations (lengths) of the object trajectories, as well as the numbers of objects in the datasets of each scene category. Some of the categories are visualized in \figurename~\ref{fig:eva_scene_categories}. Whereas \textbf{1}, \textbf{2}, \textbf{6}, \textbf{7}, and \textbf{8} show object trajectories and spatial relations, \textbf{3}, \textbf{4}, and \textbf{5} show snapshots of demonstrations. As ISM trees are generative models, different scene categories can contain the same objects and model similar relations. This also allows searching for different scenes at the same time.

\begin{table*}
\scriptsize
\centering
\caption{Performance of ISM trees learned for scene categories in our kitchen and optimized by Relation Topology Selection.}
\begin{tabular}{@{}llllllllllllll@{}}
\toprule
Scene Category & Trajectory Length & \# Objects & \multicolumn{3}{c}{Relations} & \phantom{a} & \multicolumn{3}{c}{numFPs() [\%]} & \phantom{a} & \multicolumn{3}{c}{avgDur() [s]} \\ \cmidrule{4-6} \cmidrule{8-10} \cmidrule{12-14}
&  &  & Star & Optim. & Complete && Star & Optim. & Complete && Star & Optim. & Complete \\
\midrule
Setting-Ready for Breakfast & 112 & 8 & 7 & 15 & 28 && \textcolor{cred}{38.56} & \textcolor{cdarkgreen}{3.86} & \textcolor{cdarkgreen}{0} && \textcolor{cdarkgreen}{0.044} & \textcolor{cdarkyellow}{1.228} & \textcolor{cred}{7.558} \\
Setting-Clear the Table        & 220 & 8 & 7 & 15 & 28 && \textcolor{cred}{19.59} & \textcolor{cdarkgreen}{2.92} & \textcolor{cdarkgreen}{0} && \textcolor{cdarkgreen}{0.196} & \textcolor{cdarkyellow}{2.430} & \textcolor{cred}{8.772} \\
Cupboard-Filled                   & 103 & 9 & 8 & 15 & 36 && \textcolor{cred}{23.70} & \textcolor{cdarkgreen}{8.09} & \textcolor{cdarkgreen}{0} && \textcolor{cdarkgreen}{0.020} & \textcolor{cdarkgreen}{0.140} & \textcolor{cdarkyellow}{1.070} \\
Dishwasher Basket-Filled     & 117 & 10 & 9 & 9 & 45 && \textcolor{cdarkyellow}{15.38} & \textcolor{cdarkgreen}{0} & \textcolor{cdarkgreen}{0} && \textcolor{cdarkgreen}{0.053} & \textcolor{cdarkgreen}{0.060} & \textcolor{cdarkyellow}{1.099} \\
Sandwich-Setting                 & 92 & 5 & 4 & 4 & 10 && \textcolor{cdarkgreen}{0} & \textcolor{cdarkgreen}{0} & \textcolor{cdarkgreen}{0} && \textcolor{cdarkgreen}{0.019} & \textcolor{cdarkgreen}{0.019} & \textcolor{cdarkgreen}{0.131} \\
Sandwich-on Shelf               & 98 & 6 & 5 & 7 & 15 && \textcolor{cred}{29.05} & \textcolor{cdarkgreen}{6.70} & \textcolor{cdarkgreen}{0} && \textcolor{cdarkgreen}{0.017} & \textcolor{cdarkgreen}{0.042} & \textcolor{cdarkgreen}{0.264} \\
Drinks-Setting                     & 23 & 3 & 2 & 2 & 3 && \textcolor{cred}{67.02} & \textcolor{cred}{67.02} & \textcolor{cdarkgreen}{0} && \textcolor{cdarkgreen}{0.001} & \textcolor{cdarkgreen}{0.001} & \textcolor{cdarkgreen}{0.004} \\
Drinks-on Shelf                   & 44 & 4 & 3 & 3 & 6 && \textcolor{cred}{44.51} & \textcolor{cdarkyellow}{11.54} & \textcolor{cdarkgreen}{0} && \textcolor{cdarkgreen}{0.003} & \textcolor{cdarkgreen}{0.008} & \textcolor{cdarkgreen}{0.013} \\
Cereals-Setting                   & 52 & 4 & 3 & 3 & 6 && \textcolor{cred}{33.67} & \textcolor{cdarkgreen}{0} & \textcolor{cdarkgreen}{0} && \textcolor{cdarkgreen}{0.002} & \textcolor{cdarkgreen}{0.004} & \textcolor{cdarkgreen}{0.012} \\
Cereals-on Shelf                 & 98 & 5 & 4 & 7 & 10 && \textcolor{cdarkyellow}{9.50} & \textcolor{cdarkgreen}{1.68} & \textcolor{cdarkgreen}{0} && \textcolor{cdarkgreen}{0.014} & \textcolor{cdarkgreen}{0.053} & \textcolor{cdarkgreen}{0.181} \\
\bottomrule
\end{tabular}
\label{tab:eva_ism_tree_performances}
\end{table*}

In the different areas of our setup, the objects can be arranged horizontally or vertically in 2D. However, we define scene categories that span multiple areas and thus extend into 3D. For instance, the category ``Cereals-on Shelf'' in \textbf{5}, \textbf{6}, and \textbf{8} relates parts of a table setting to the food and drinks stored on the shelves. Food, drinks, and the shelves are also part of ``Drinks-on Shelf'' in \textbf{2} and ``Sandwich-on Shelf''. The ISM trees for these scene categories contain relations of a considerable length, such as those drawn in \textbf{2}. The object configurations corresponding to these scene categories are truly three-dimensional, as they extend both horizontally and vertically. A close-up view of the vertical relations in ``Cereals-on Shelf'' is provided in \textbf{6}, whereas the horizontal ones are shown in \textbf{8}. Except for the shelves, the three categories ``Sandwich-Setting'' in \textbf{1} and \textbf{3}, ``Cereals-Setting'' in \textbf{4} and \textbf{7}, and ``Drinks-Setting'' consist of the same objects as their ``...-on Shelf'' counterparts. The former three expect food and drinks to be located on a table, not on the shelves.

\subsubsection{Performance of Optimized ISM Trees}\label{sec:ear_eorts_pooit}

\begin{figure}[h!]
  \centering
      \includegraphics[width=0.8\linewidth]{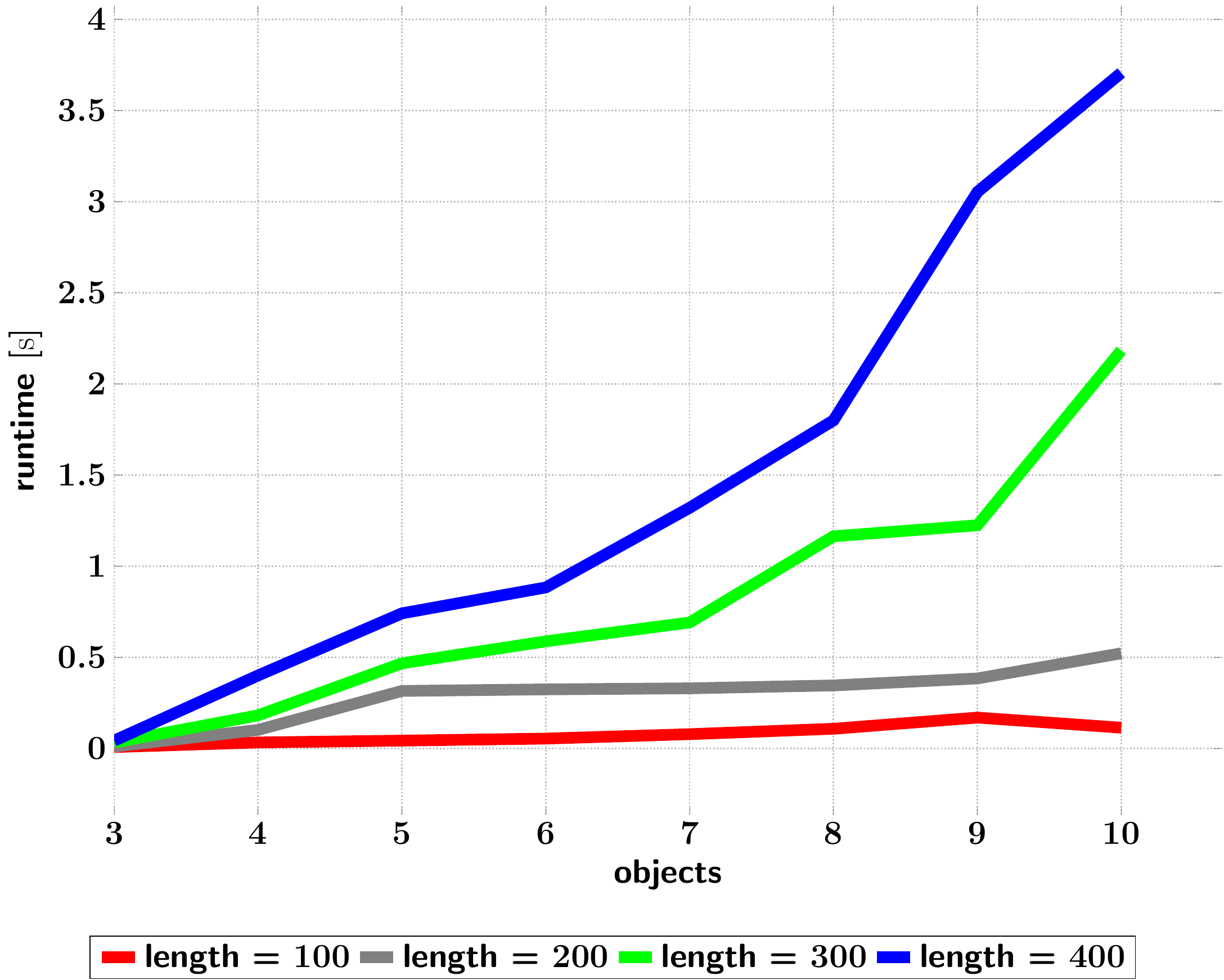}
    \caption{Times which ISM trees (from optimized topologies) take to recognize scenes, depending on the trajectory length in their datasets.}
  \label{fig:eva_psr_runtimes}
\end{figure}

\tablename~\ref{tab:eva_ism_tree_performances} shows how the ISM trees for the scene categories from the previous subsection perform concerning the goodness measures numFPs\(()\) (given in percent) and avgDur\(()\). One line corresponds to one category. The two measures were defined in Sec. \ref{sec:rts_co} for the RTS. We rate their values with color coding, as in Sec. \ref{sec:ear_eopsr_pdsci}. Additionally, the table specifies the trajectory lengths and numbers of objects for each category and the number of relations it models. The presented values confirm that the runtime of scene recognition with a complete topology is orders of magnitude higher than that with a star. Especially for the larger categories in \tablename~\ref{tab:eva_ism_tree_performances}, complete topologies are much too inefficient for ASR. It should be noted that high runtimes do not only result from large numbers of objects, but also from long demonstration recordings. This, e.g., explains the runtime difference between ``Setting-Ready for Breakfast'' and ``Cupboard-Filled''. The table also displays the high numbers of false positives produced with star topologies, so that they are not an alternative to complete topologies. However, the number of false positives also depends on how much the objects in a category have been moved during a demonstration. If the objects are barely moved, such as for ``Sandwich-Setting'', a star topology is just as reliable as a complete topology. Overall, however, only optimized topologies achieve simultaneously low values for numFPs\(()\) and for avgDur\(()\).

\begin{table*}[h!]
\scriptsize
\centering
\caption{Performance measures for experiments with the physical MILD robot, used to evaluate Active Scene Recognition.}
\begin{tabular}{@{}llllllllllllll@{}}
\toprule
Task & Duration [s] & Camera Views & Found Objects [\%] & \multicolumn{10}{c}{Confidences} \\ \cmidrule{5-14}
 &  &  &  & Setting & \phantom{a} & \multicolumn{2}{c}{Drinks} & \phantom{a} & \multicolumn{2}{c}{Cereals} & \phantom{a} & \multicolumn{2}{c}{Sandwich} \\ \cmidrule{5-5} \cmidrule{7-8} \cmidrule{10-11} \cmidrule{13-14}
 &  &  &  & Ready && Setting & on Shelf && Setting & on Shelf && Setting & on Shelf \\
\midrule
s1\_e1 & \textcolor{cdarkyellow}{783.96} & \textcolor{cdarkyellow}{16} & \textcolor{cdarkgreen}{100} & \textcolor{cdarkgreen}{0.97} && \textcolor{cdarkgreen}{0.33} & \textcolor{cdarkgreen}{0.99} && \textcolor{cdarkgreen}{0.49} & \textcolor{cdarkyellow}{0.85} && \textcolor{cdarkgreen}{0.99} & \textcolor{cdarkgreen}{0.34} \\
             & \textcolor{cdarkgreen}{425.63} & \textcolor{cdarkgreen}{9}  & \textcolor{cdarkgreen}{100} & \textcolor{cdarkgreen}{0.99} && \textcolor{cdarkgreen}{0.47} & \textcolor{cdarkgreen}{0.97} && \textcolor{cdarkgreen}{0.48} & \textcolor{cdarkgreen}{0.99} && \textcolor{cdarkgreen}{1.00} & \textcolor{cdarkgreen}{0.37} \\
s1\_e2 & \textcolor{cdarkgreen}{560.18} & \textcolor{cdarkgreen}{12} & \textcolor{cdarkgreen}{100} & \textcolor{cdarkgreen}{0.97} && \textcolor{cdarkgreen}{0.33} & \textcolor{cdarkyellow}{0.42} && \textcolor{cdarkgreen}{0.46} & \textcolor{cdarkyellow}{0.82} && \textcolor{cdarkgreen}{0.99} & \textcolor{cdarkgreen}{0.49} \\
             & \textcolor{cdarkgreen}{562.46} & \textcolor{cdarkgreen}{14} & \textcolor{cdarkgreen}{100} & \textcolor{cdarkgreen}{0.97} && \textcolor{cdarkgreen}{0.33} & \textcolor{cdarkyellow}{0.41} && \textcolor{cdarkgreen}{0.47} & \textcolor{cdarkyellow}{0.89} && \textcolor{cdarkgreen}{0.99} & \textcolor{cdarkgreen}{0.49} \\
s2\_e1 & \textcolor{cdarkgreen}{367.32} & \textcolor{cdarkgreen}{8}  & \textcolor{cdarkgreen}{100} & \textcolor{cdarkgreen}{0.93} && \textcolor{cdarkgreen}{0.33} & \textcolor{cdarkgreen}{0.98} && \textcolor{cdarkgreen}{0.96} & \textcolor{cdarkgreen}{0.75} && \textcolor{cdarkgreen}{1.00} & \textcolor{cdarkgreen}{0.45} \\
             & \textcolor{cdarkgreen}{336.18} & \textcolor{cdarkgreen}{8}  & \textcolor{cdarkgreen}{100} & \textcolor{cdarkgreen}{0.91} && \textcolor{cdarkgreen}{0.33} & \textcolor{cdarkgreen}{0.98} && \textcolor{cdarkgreen}{0.96} & \textcolor{cdarkgreen}{0.74} && \textcolor{cdarkgreen}{0.99} & \textcolor{cdarkgreen}{0.50} \\
s2\_e2 & \textcolor{cdarkgreen}{584.07} & \textcolor{cdarkgreen}{13} & \textcolor{cdarkgreen}{100} & \textcolor{cdarkgreen}{0.95} && \textcolor{cdarkgreen}{0.33} & \textcolor{cdarkgreen}{0.73} && \textcolor{cdarkgreen}{0.97} & \textcolor{cdarkgreen}{0.68} && \textcolor{cdarkgreen}{0.99} & \textcolor{cdarkgreen}{0.34} \\
             & \textcolor{cdarkgreen}{434.99} & \textcolor{cdarkgreen}{10} & \textcolor{cdarkgreen}{100} & \textcolor{cdarkgreen}{0.92} && \textcolor{cdarkgreen}{0.33} & \textcolor{cdarkgreen}{0.75} && \textcolor{cdarkgreen}{0.98} & \textcolor{cdarkgreen}{0.69} && \textcolor{cdarkgreen}{1.00} & \textcolor{cdarkgreen}{0.42} \\
s3\_e1 & \textcolor{cdarkgreen}{533.29} & \textcolor{cdarkgreen}{11} & \textcolor{cdarkgreen}{93.75} & \textcolor{cdarkgreen}{0.99} && \textcolor{cdarkyellow}{0.93} & \textcolor{cdarkgreen}{0.25} && \textcolor{cdarkgreen}{0.99} & \textcolor{cdarkgreen}{0.50} && \textcolor{cdarkgreen}{1.00} & \textcolor{cdarkgreen}{0.34} \\
             & \textcolor{cdarkgreen}{439.61} & \textcolor{cdarkgreen}{8}  & \textcolor{cdarkgreen}{100} & \textcolor{cdarkgreen}{0.99} && \textcolor{cdarkyellow}{0.91} & \textcolor{cdarkgreen}{0.48} && \textcolor{cdarkgreen}{0.98} & \textcolor{cdarkyellow}{0.75} && \textcolor{cdarkgreen}{1.00} & \textcolor{cdarkgreen}{0.37} \\
\bottomrule
\end{tabular}
\label{tab:eva_asr_values_s2}
\end{table*}

\begin{figure*}[h!]
  \centering
      \includegraphics[width=0.71\linewidth]{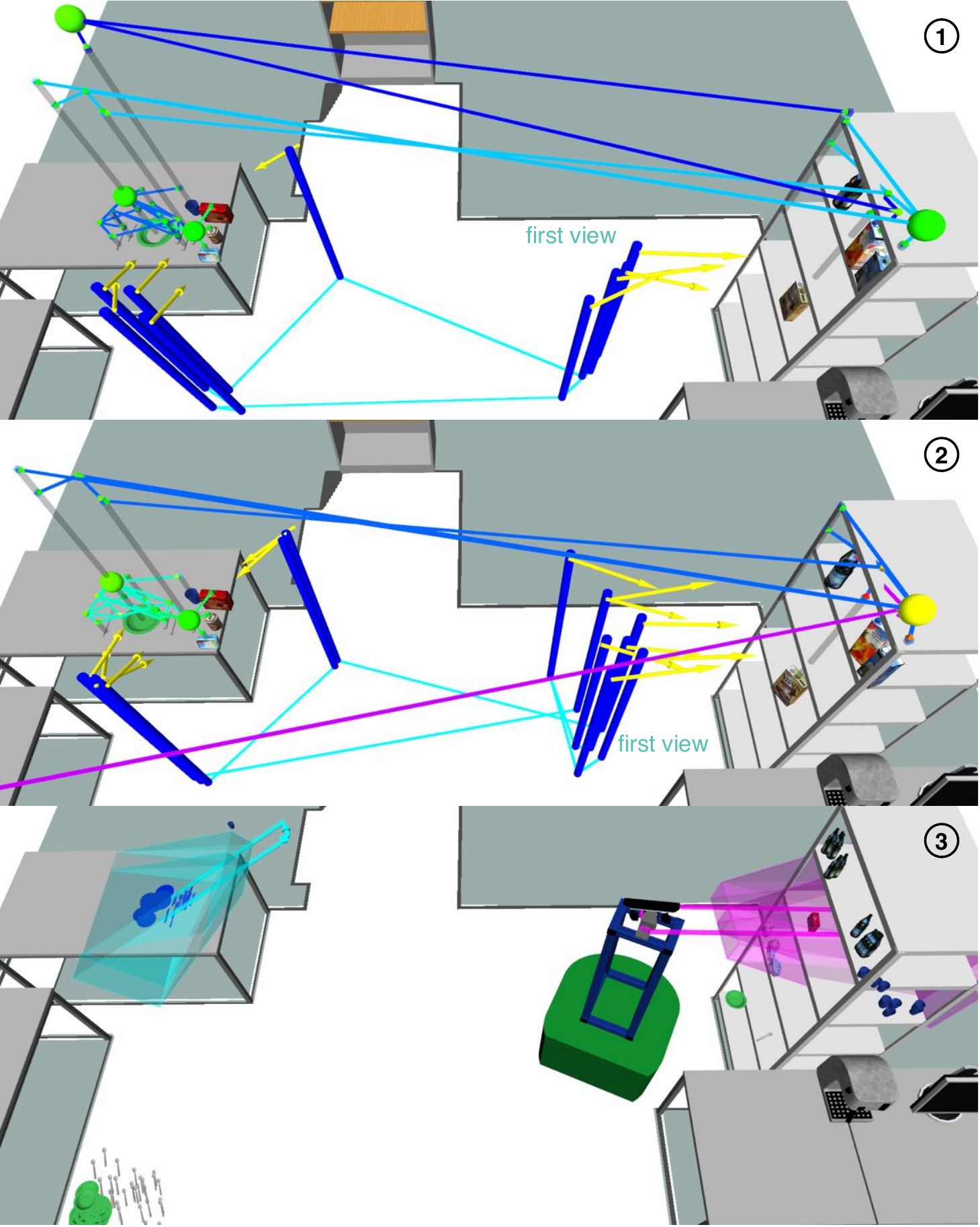}
  \caption{Influence of object orientations on pose prediction: \textbf{1} and \textbf{2} show s1\_e1 and s1\_e2. Between s1\_e1 and s1\_e2, all objects on the shelves (right) were rotated. \textbf{1}, \textbf{2}: Recognized scenes and camera views MILD adopted. \textbf{3}: Snapshot during the execution of s1\_e2. The miniature objects correspond to predicted poses.}
  \label{fig:eva_results_s2_m1}
\end{figure*}

In \figurename~\ref{fig:eva_psr_runtimes}, we now consider the runtime of scene recognition with ISM trees from optimized topologies individually. This plot shows average runtimes for different datasets, depending on the number of objects they contain and the length of the record of their demonstration. Unlike \tablename~\ref{tab:eva_ism_tree_performances} which presents results from sensor-recorded demonstrations, all object trajectories for this plot are generated in simulation. Scene recognition is performed on 600 object configurations generated for each dataset in \figurename~\ref{fig:eva_psr_runtimes}. The recognition runtimes for these configurations are given in seconds and each curve stands for a specific trajectory length. All curves appear to be linear for the number of objects. The slopes of the curves appear to be determined by the trajectory lengths. Thus, the runtimes appear to correlate with a product of trajectory length and number of objects. Beyond such favorable time complexity, another experiment renders ISM trees suitable for object search applications: We measured a maximum runtime of 3.71 seconds for ten objects and a trajectory length of 400 samples (20 min when capturing samples every 3 seconds while a demonstration is recorded). The fact that the runtimes with optimized topologies in \tablename~\ref{tab:eva_ism_tree_performances} remain under 2.5 seconds further emphasizes that ISM trees are suitable for the state machine we use to implement ASR.

\subsection{Evaluation of Active Scene Recognition}\label{sec:ear_eoasr}

\begin{figure*}[h!]
  \centering
      \includegraphics[width=0.71\linewidth]{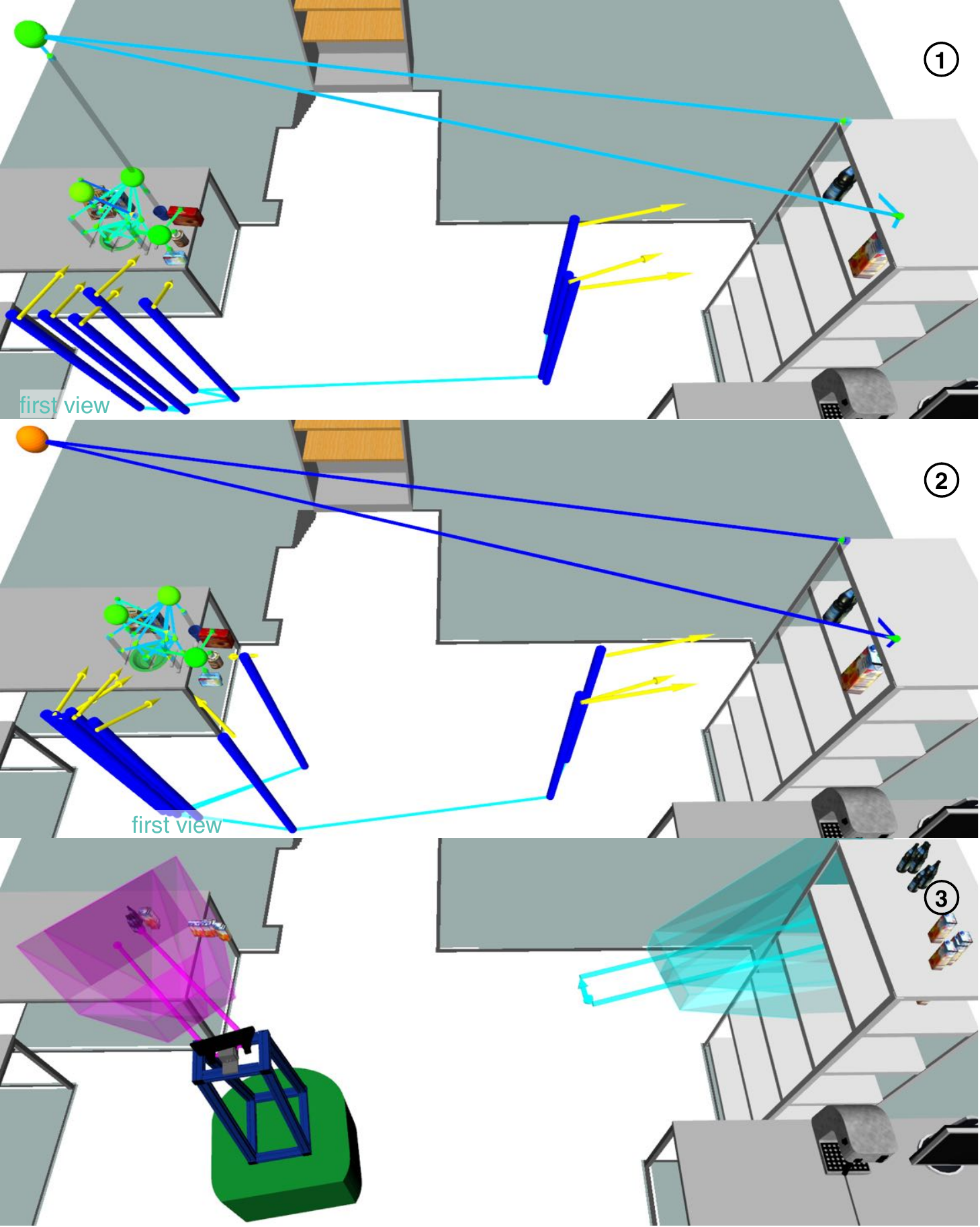}
  \caption{Influence of object positions on pose prediction: \textbf{1} and \textbf{2} show s2\_e1 and s2\_e2. Between s2\_e1 and s2\_e2, all objects on the table were shifted to the right. \textbf{1}, \textbf{2}: Recognized scenes and camera views MILD adopted. \textbf{3}: Snapshot during the execution of s2\_e2. The miniature objects correspond to predicted poses.}
  \label{fig:eva_results_s2_m3}
\end{figure*}

\subsubsection{Influence of Object Orientations on Pose Prediction} \label{sec:eva_evaasr_roswodr}

Unlike Sec. \ref{sec:ear_eopsr_ioop}, in this section, we no longer detect objects from a single viewpoint. Here, we investigate how well our ASR approach can recognize scenes whose object configurations cannot be fully perceived from a single viewpoint. We evaluated ASR in our kitchen setup in six experiments. Each experiment was performed twice to account for the positioning uncertainties of our MILD robot. In these experiments, the robot was expected to recognize all existing instances of the scene categories specified in \tablename~\ref{tab:eva_asr_values_s2}. The experiments in Sec. \ref{sec:eva_evaasr_roswodr} and \ref{sec:eva_evaasr_roswodt} analyze how our object pose prediction, and thus ASR, is affected when we change object poses between the previous demonstration and the execution of ASR during the experiment. Sec. \ref{sec:eva_evaasr_roswodr} focuses on rotational displacements, whereas Sec. \ref{sec:eva_evaasr_roswodt} addresses translational displacements.

In both subsections, we investigate how well ASR can detect all objects in scene category instances despite such displacements and whether it can detect the deviations from learned relations that result from these displacements. The first experiments (s1\_e1 and s2\_e1) in each subsection are performed on object configurations identical to the demonstration, whereas the second experiments (s1\_e2 and s2\_e2) cover either rotational or translational displacements. At the beginning of s1\_e1 and s1\_e2, MILD looks at the upper right corner of the shelves in \textbf{1} in \figurename~\ref{fig:eva_results_s2_m1}. \textbf{1} visualizes the results of one of the two executions of s1\_e1 in \tablename~\ref{tab:eva_asr_values_s2}. From there, MILD detects two searched objects. It then predicts object poses on the shelves and the table. NBV estimation minimizes the travel time for MILD by letting ASR search for another object on the shelves. Video clip 3 (``Influence of Object Orientations on Pose Prediction'') shows how MILD proceeds further. In the end, instances of the categories ``Drinks-on Shelf'', ``Cereals-on Shelf'', ``Sandwich-Setting'', and ``Setting-Ready for Breakfast'' are correctly recognized.

As visible in \textbf{2}, all searched objects on shelves were rotated for s1\_e2. This change affects the categories ``Drinks-on Shelf'' and ``Cereals-on Shelf''. That the confidence levels for the two categories fall off differently seems counterintuitive, but is because they do not contain the same number of objects. s1\_e2 also shows that ASR requires only a single correctly oriented object (here, the shelves) to find all objects from the same scene category. This is because the shelves cause correctly predicted poses on the table, so ASR can ignore the more distant incorrect predictions at the lower left of \textbf{3} that result from the rotated objects. Which views the robot has just adopted and will adopt next is visualized by red and turquoise frustums. Predicted poses that are within the future view are colored blue. The large gap between correct and incorrect pose predictions illustrates the extent to which rotational changes can affect pose prediction when using long spatial relations.

\subsubsection{Influence of Object Positions on Pose Prediction} \label{sec:eva_evaasr_roswodt}

At the beginning of s2\_e1, MILD is not standing in front of the shelves, but in front of the table. As shown in \textbf{1} in \figurename~\ref{fig:eva_results_s2_m3}, MILD first searches this table. We have recorded in video clip 4 (``Influence of Object Positions on Pose Prediction'') how MILD proceeds until all existing scene categories are recognized. The difference between s2\_e1 and s2\_e2 is that all objects on the table were shifted at the same time using a tray. The confidence levels of those categories in \tablename~\ref{tab:eva_asr_values_s2}, which contain only objects on the table, remain unchanged. Since all their objects are on the tray, shifting the tray does not affect the relative poses between them. This shows that ISM trees depend only on relative object poses and not directly on absolute object poses, thanks to the relations used.

The long relations between the objects on the table and on the shelves are affected by the shift, but only slightly. The predicted poses on the shelves move backwards, as shown in \textbf{2} in \figurename~\ref{fig:eva_results_s2_m3}. However, they stay close enough to the shelves, so MILD still finds all objects. Yet, even a small orientation error in an object estimate on the table causes some predicted poses on one shelf to move up one level, as shown on the right in \textbf{3}. Overall, the effect of rotational deviations on the accuracy of pose prediction depends on the length of the relation used, while that of translational deviations is constant.

\subsubsection{Active Scene Recognition on a Cluttered Table}\label{sec:ear_eoasr_rosiacoc}

\begin{figure}[tpb]
  \centering
      \includegraphics[width=0.85\linewidth]{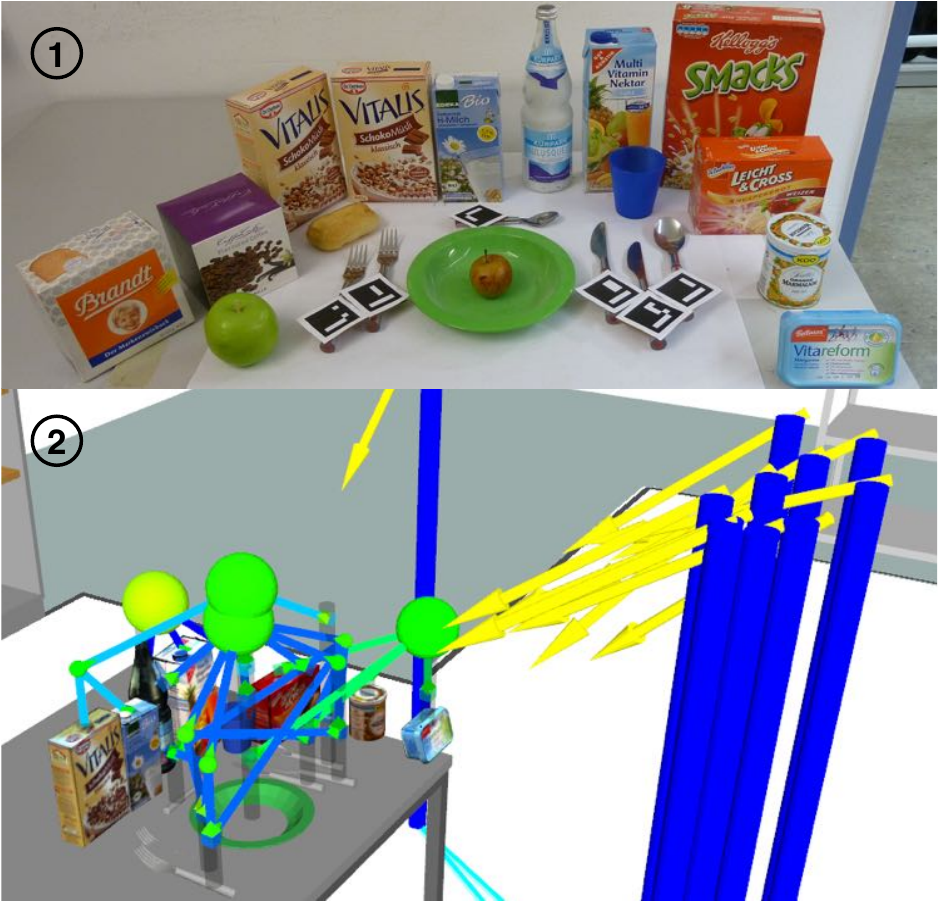}
  \caption{Active Scene Recognition on a cluttered table. \textbf{1}: Snapshot of physical objects. \textbf{2}: Camera views adopted and recognized scene instances.}
  \label{fig:eva_s2_m4}
\end{figure}

After two subsections devoted to object configurations spread across our kitchen setup, this subsection shows how our ASR approach deals with an object configuration that brings together a large number of searched objects from different overlapping scenes. Such a configuration - a cluttered table - can be seen in \textbf{1} in \figurename~\ref{fig:eva_s2_m4}. It consists of 15 objects to be searched, several of which are obscured from certain viewpoints, and seven irrelevant objects. As \textbf{2} in \figurename~\ref{fig:eva_s2_m4} and video clip 5 (``Active Scene Recognition on a Cluttered Table'') show, MILD manages to find all searched objects. The only object that is not always found in the corresponding experiment s3\_e1 is the shelves and these do not participate in s3\_e1. The objects at the front of the table are easily localized, and scene recognition achieves high confidence levels, e.g., for ``Setting-Ready for Breakfast''. The objects in the back are more difficult to find, resulting in a lower confidence level for ``Drinks-Setting''. The spuriously high confidence level for ``Cereals-on Shelf'' results from a false positive returned by the object localization. All irrelevant objects are correctly discarded by ASR.

\begin{figure*}[h!]
  \centering
      \includegraphics[width=\linewidth]{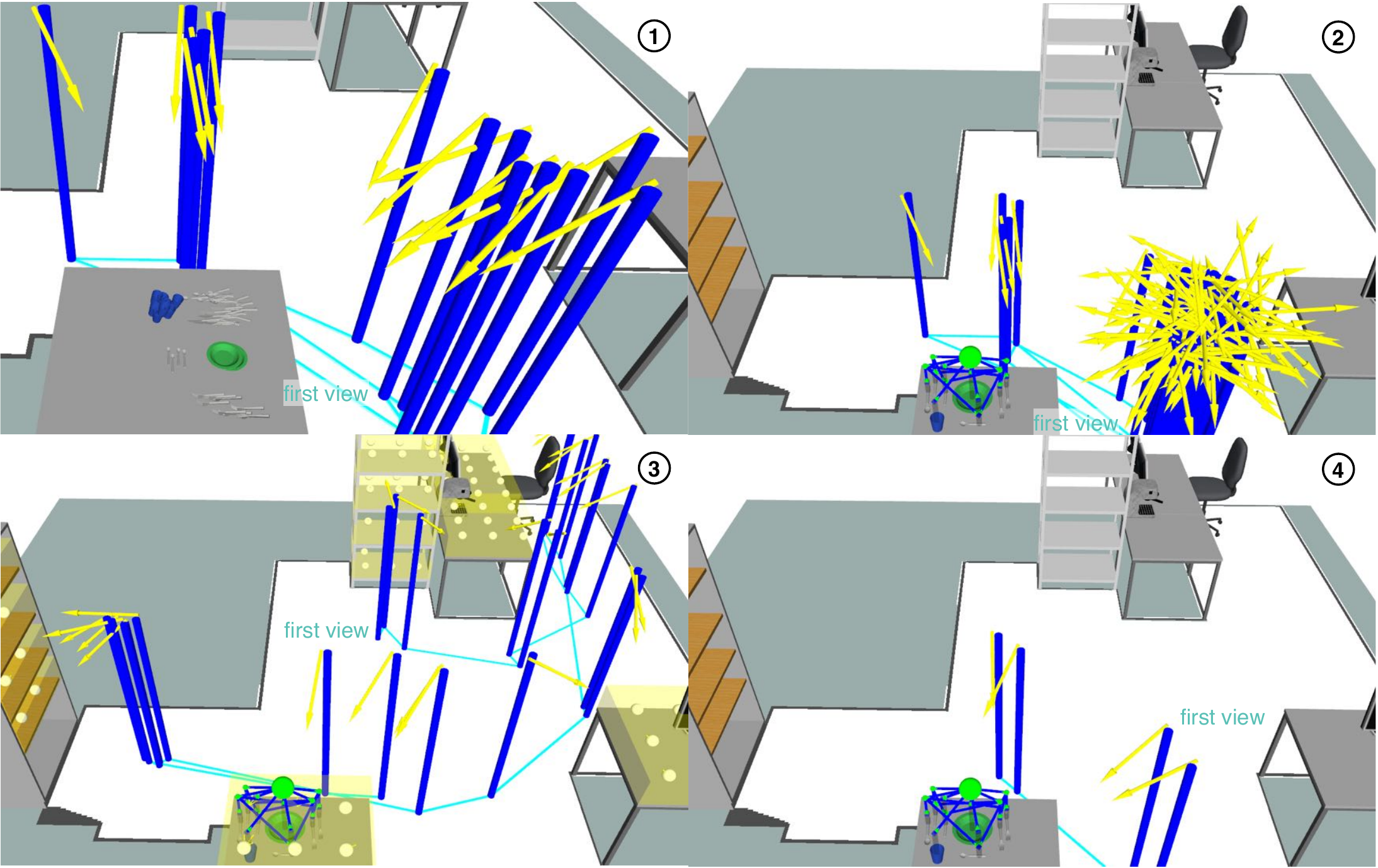}
      \caption{Comparison of three approaches to ASR: For ``direct search only'' (see \textbf{1},\textbf{2}), ``bounding box search'' (see \textbf{3}), and our ASR approach (see \textbf{4}), we show which camera views were adopted. Additionally, \textbf{1} shows demonstrated object poses and \textbf{2}-\textbf{4} show recognized scene instances.}
  \label{fig:eva_comparison}
\end{figure*}

\subsubsection{Comparison of Three Approaches to ASR}\label{sec:ear_eoasr_eocotata}

In this subsection, we compare the execution times of our approach to ASR to those achieved by two alternative approaches to ASR. The searched object configuration looks similar to ``Setting-Ready for Breakfast''. The first alternative to our approach is called ``direct search only'' and omits the INDIRECT\_SEARCH mode. Instead, its SCENE\_RECOGNITION substate exclusively processes object estimates acquired by the informed and uninformed strategies of the DIRECT\_SEARCH mode. We call the second alternative ``bounding box search''. This approach assumes that objects can only be located in so-called bounding boxes determined by prior knowledge and does not use INDIRECT\_SEARCH to predict the poses of searched objects. Such bounding boxes are visualized in \textbf{3} in \figurename~\ref{fig:eva_comparison} as yellow boxes in which possible object poses are visualized as white spheres. However, ``bounding box search'' uses Nest-Best-Views (NBVs) to sweep the bounding boxes.

Compared to the previous experiments (see \textbf{1} in \figurename~\ref{fig:eva_s2_m4}), the searched place setting is shifted and rotated (see \textbf{2} - \textbf{4} in \figurename~\ref{fig:eva_comparison}). All three ASR approaches are executed twice and successfully find all objects in the setting. Since ``direct search only'' and ``bounding box search'' take an inordinate amount of time, we ran this experiment in simulation. Both alternatives take much longer than our ASR approach: 31.13 and 15.24 minutes instead of 2.48 minutes. The informed strategy of ``direct search only'' is not able to find all objects. The views the strategy adopts are shown in \textbf{1} in \figurename~\ref{fig:eva_comparison}. MILD then uses the uninformed strategy of ``direct search only'', which causes a lengthy search but eventually succeeds. \textbf{2} shows the views adopted by both the informed and uninformed strategies. The views in \textbf{3} are adopted by ``bounding box search'' and show that this approach often moves between bounding boxes rather than searching a single one from different perspectives. This highlights the difficulty of parameterizing the estimation of NBVs. In \textbf{4}, our approach adopts the same first two views as ``direct search only'' in \textbf{1}. However, instead of continuing the search on the right of the table once some objects are found, our approach adapts to the fact that the place setting has been rotated and lets MILD search on the other side of the table.

\subsubsection{Runtime of Object Pose Prediction}\label{sec:ear_eoasr_r}

\begin{figure}[h!]
  \centering
      \includegraphics[width=0.8\linewidth]{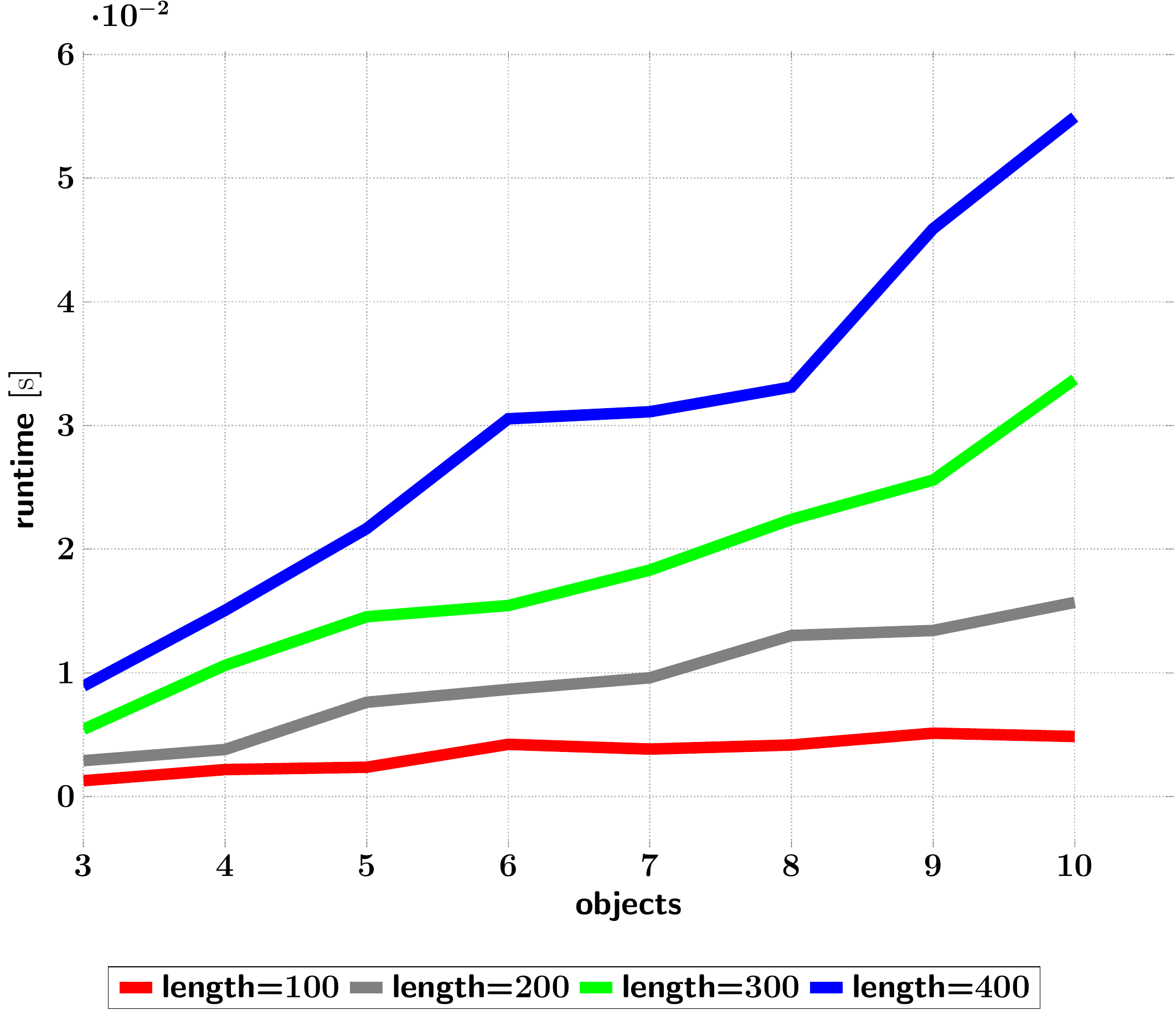}
    \caption{Times which our pose prediction algorithm takes, depending on the number of objects and trajectory length in the datasets.}
  \label{fig:eva_opp_runtimes}
\end{figure}

We reused the ISM trees from Sec. \ref{sec:ear_eorts_pooit} to compute the runtime of our pose prediction algorithm for datasets with different numbers of objects \(n\) and trajectory lengths \(l\). We averaged \(10 \cdot n \cdot l\) executions of the algorithm per dataset in \figurename~\ref{fig:eva_opp_runtimes}. Each runtime shown corresponds to the time required to predict the poses of all objects in a category. If we disregard scaling (runtimes are given here in hundredths of a second), the analysis of the curves from Sec. \ref{sec:ear_eorts_pooit} also applies to \figurename~\ref{fig:eva_opp_runtimes}. Given that the maximum runtime is 0.055 seconds for ten objects and a trajectory length of 400 samples, the time consumption of object pose prediction seems negligible compared to the one of scene recognition.

\section{Conclusions}\label{sec:conclusionsandfutureworks}

Through its contributions, this article closes three gaps in the core of Active Scene Recognition (ASR). ASR, which was impractical without these contributions, combines scene recognition and object search - two tasks that are otherwise considered separately. Firstly, ASR enables scene recognition to analyze object configurations that cannot be perceived from a single viewpoint. Secondly, it allows object search to be guided by object configurations rather than single objects, making it more efficient. Using only single objects can lead to ambiguities, since, e.g., a knife in a table setting would expect a plate to be beneath itself when a meal is finished, while it would expect the plate to be beside itself when the meal has not yet started.

The feature extraction components of part-based models may be outdated compared to Convolutional Neural Nets (today's gold standard). However, this article aims to show that ISMs are nevertheless particularly suitable for modeling the spatial characteristics of relations and their uncertainties. ISM trees additionally overcome the limitation of single ISMs to represent only a single relation topology. Replacing the feature extraction of ISMs with appropriate object pose estimators, ISM trees provide up-to-date object-based scene classification. Therefore, they can be seen as a complement to Convolutional Neural Nets. Especially when modeling relations in scenes that express personal preferences and for which only small amounts of data are available, a technique such as ISMs is suitable. This suitability stems from the fact that ISMs model relations nonparametrically in the sense of instance-based learning (\cite{mitchell1997machine}).

However, the fact that ISM trees model relations nonparametrically also means that they can be prone to combinatorial explosion. To avoid such effects in the recognition and prediction algorithms we contribute, we have implemented the following strategies: As proposed by \cite{leibe2008robust}, an accumulator array and a method similar to Mean-Shift Search are used within single ISMs during recognition to prune a significant portion of the votes. Moreover, when recognizing scenes with an ISM tree, two factors - the number of intermediate results passed from one ISM to the next and the lengths of the chains of interrelated ISMs in a tree - can cause combinatorial effects. We limit the first through passing only the best-rated intermediate results between ISMs and the second by minimizing the heights of ISM trees through our tree generation algorithm. Moreover, we solve the combinatorial explosion that made our previous pose prediction algorithm inefficient. Instead of simply concatenating inverted spatial relations in an ISM tree, the new method samples random subsets from these relations.

Our evaluation of PSR in Sec. \ref{sec:ear_eopsr} provided evidence that any ISM of an ISM tree precisely detects when object poses deviate from modeled spatial relations. Depending on their parametrization, ISMs are more or less permissive concerning such deviations. Further experiments in \cite{meissner2018} have also shown that ISM trees are robust against objects missing in object configurations. In Sec. \ref{sec:ear_eoasr}, we applied ASR onto object configurations which were considerably more complex than those used in our previous work. Robot localization and object pose estimation accuracy were the limiting factors for our ASR approach. Still, ASR even succeeded in recognizing scenes at different locations with the same ISM tree. This illustrates that spatial relations make ISM trees particularly reusable compared to techniques that model scenes using absolute object poses. Video clip 6 (``Recognition of scenes independent of their locations'') is devoted to this major advantage of ISM trees and thus ASR. The experiments in Sec. \ref{sec:ear_eorts_pooit} suggest that the runtime of PSR linearly depends on the number of objects included in a scene category. An experiment we conducted for datasets including six objects indicates that this is also true for trajectory lengths: We measured a maximum recognition runtime of 4.94 seconds for a demonstration trajectory length of 1000 samples. Still, the recognition runtimes for longer trajectory lengths can exceed the requirements of ASR. To overcome this limitation, we plan to compress the relations in ISM trees by eliminating redundant relative poses with downsampling voxel grids.

\appendix

\section{Acknowledgments}

This work draws from student's projects by Fabian Hanselmann, Heinreich Heizmann, Oliver Karrenbauer, Felix Marek, Jonas Mehlhaus, Patrick St\"ockle, Reno Reckling, Daniel Stroh, and Jeremias Trautmann. Our special thanks go to Rainer J\"akel, Michael Beetz, and Torsten Kr\"oger for their valuable advice.

\section{Pseudocodes for Contributions}

This subsection provides pseudocodes for the three contributions of this article. Algo. \ref{alg:psr_generateISMTreeFromSubTopologies} is our algorithm for generating ISM trees and corresponds to contribution 1. Algo. \ref{alg:psr_evaluateISMsInTree} to \ref{alg:psr_findSubInstances} form our algorithm for scene recognition using ISM trees and correspond to contribution 2. Algo. \ref{alg:asr_predictPose} and \ref{alg:asr_generateCloudOfPosePredictions} form our algorithm for predicting object poses with ISM trees and correspond to contribution 3. As it is not the goal of this article to describe all the details of these algorithms, but to present their key ideas concisely, some variables and helper functions are defined only in \cite{meissner2018}.

\begin{minipage}{\linewidth}
\begin{algorithm}[H]
\begin{algorithmic}[1]
{\small
\caption{generateISMTreeFromStarTopologies$(z, \{\Sigma_{\sigma}\},$ $\{\textbf{J}(o)\}) \rightarrow \{m\}$.}
\label{alg:psr_generateISMTreeFromSubTopologies}
\STATE $i \gets |\{\Sigma_{\sigma}\}| - 2$ and parentFound $\gets$ \FALSE
\STATE $\displaystyle h_{\{\Sigma_{\sigma}\}} \gets \max_{o \in \{o\}} h_{\{\Sigma_{\sigma}\}}(o)$
\FOR{$h_{j} \gets h_{\{\Sigma_{\sigma}\}},\dots,0$}
\FORALL{$\Sigma_{\sigma}(j) \in \left\{ \Sigma_{\sigma}(j) \left| \Sigma_{\sigma}(j) \in \{\Sigma_{\sigma}\} \wedge \exists o \in \right. \right.$ $\left. \left. H_{D}(\Sigma_{\sigma}(j)): h_{\{\Sigma_{\sigma}\}}(o) = h_{j} \right. \right\}$}
\STATE Randomly extract $o_{H}$ from $\argmin\limits_{o \in H_{D}(\Sigma_{\sigma}(j))}\, h_{\{\Sigma_{\sigma}\}}(o)$
\IF{$h_{j} = 0$}
\STATE $z_{m} \gets z$
\ELSE
\STATE $z_{m} \gets $ append(append($z$,''\_sub''), $i$)
\STATE $i \gets i - 1$
\ENDIF
\STATE $\left(m, \textbf{J}_F\right) \gets$ learnISM$(z_{m}, \textbf{J}(o_{M}(\Sigma_{\sigma}(j))) \cup \left\{ \textbf{J}(o) \left| \right. \right.$  \\ $\left.\textbf{J}(o) \in \{\textbf{J}(o)\} \wedge o \in N(o_{M}(\Sigma_{\sigma}(j))) \right\})$
\STATE Create $o_{F}$ with $\textbf{J}_F$ as its trajectory
\STATE $\{m\} \gets \{m\} \cup m$ and $\{\Sigma_{\sigma}\} \gets \{\Sigma_{\sigma}\} \setminus \Sigma_{\sigma}(j)$
\FOR{$h_{k} \gets 0,\dots,h_{j}-1$}
\FORALL{$\Sigma_{\sigma}(k) \in \left\{ \Sigma_{\sigma}(k) \left| \Sigma_{\sigma}(k) \in \{\Sigma_{\sigma}\} \wedge \exists o \in \right. \right.$ $\left. \left. H_{D}(\Sigma_{\sigma}(k)): h_{\{\Sigma_{\sigma}\}}(o) = h_{k} \right. \right\}$}
\IF{$o_{H} \in N(o_{M}(\Sigma_{\sigma}(k)))$}
\STATE $N(o_{M}(\Sigma_{\sigma}(k))) \gets (N(o_{M}(\Sigma_{\sigma}(k))) \setminus o_{H}) \cup o_{F}$
\STATE $\{\textbf{J}(o)\} \gets \{\textbf{J}(o)\} \cup \textbf{J}_F$ and $h_{\{m\}}(m) \gets h_{k}$
\STATE parentFound $\gets$ \TRUE
\BREAK
\ENDIF
\ENDFOR
\IF{parentFound = \TRUE}
\STATE parentFound $\gets$ \FALSE
\BREAK
\ENDIF
\ENDFOR
\ENDFOR
\ENDFOR
\RETURN{\{m\}}
}
\end{algorithmic}
\end{algorithm}

\begin{algorithm}[H]
\begin{algorithmic}[1]
{\small
\caption{evaluateISMsInTree$(\{i\}, \{m\}) \rightarrow \{\textbf{I}_{\{m\}}\}$.}
\label{alg:psr_evaluateISMsInTree}
\FOR{$h \gets h_{\{m\}},\dots,1$}
\FORALL{$\left\{ m \left| m \in \{m\} \wedge h_{\{m\}}(m) = h \right. \right\}$}
\STATE $\{\textbf{I}_{m}\} \gets$ recognitionSingleISM$(\{i\}, m)$
\FORALL{$\textbf{I}_{m} \in \{\textbf{I}_{m}\}$}
\STATE Create $o_{F}$ with $\textbf{E}(o_{F}) = (z,0,\textbf{T}_{F})$ and $b(o_{F}) = b_{F}$, all extracted from $\textbf{I}_{m}$
\STATE $\{i\} = \{i\} \cup \{o_{F}\}$
\ENDFOR
\STATE $\{\textbf{I}_{\{m\}}\} = \{\textbf{I}_{\{m\}}\} \cup \{\textbf{I}_{m}\}$
\ENDFOR
\ENDFOR
\STATE $\{\textbf{I}_{m_{R}}\} \gets$ recognitionSingleISM$(\{i\}, m)$
\STATE $\{\textbf{I}_{\{m\}}\} = \{\textbf{I}_{\{m\}}\} \cup \{\textbf{I}_{m_{R}}\}$
\RETURN{$\{\textbf{I}_{\{m\}}\}$}
}
\end{algorithmic}
\end{algorithm}

\begin{algorithm}[H]
\begin{algorithmic}[1]
{\small
\caption{assembleInstances$(\{\textbf{I}_{\{m\}}\}, \epsilon_{R})$ $\rightarrow \{\textbf{I}_{\textbf{S}}\}$.}
\label{alg:psr_assembleSceneCategoryInstances}
 \FORALL{$\textbf{I}_{m_{R}} \in \{\textbf{I}_{\{m\}}\}$}
 \IF{$b(\textbf{I}_{m_{R}}) \geq \epsilon_{R}$}
 \STATE $\{\textbf{I}\} \gets$ findSubInstances$(\textbf{I}_{m_{R}}, \{\textbf{I}_{\{m\}}\})$
 \ENDIF
\STATE $\textbf{I}_{\textbf{S}} \gets \{\textbf{I}\} \cup \textbf{I}_{m_{R}}$
\STATE $\{\textbf{I}_{\textbf{S}}\} \gets \{\textbf{I}_{\textbf{S}}\} \cup \textbf{I}_{\textbf{S}}$
 \ENDFOR
\RETURN{$\{\textbf{I}_{\textbf{S}}\}$}
}
\end{algorithmic}
\end{algorithm}
\end{minipage}

\begin{minipage} {\linewidth}
\begin{algorithm}[H]
\begin{algorithmic}[1]
{\small
\caption{findSubInstances$(\textbf{I}_{m'}, \{\textbf{I}_{\{m\}}\}) \rightarrow \{\textbf{I}\}$.}
\label{alg:psr_findSubInstances}
 \STATE Extract $ \{i\}_{m'} $ from $ \textbf{I}_{m'}$
 \FORALL{$i \in \{i\}_{m'} $}
 \STATE Extract $c$ and $\textbf{T}$ from $\textbf{E}(i)$
 \FORALL{$\textbf{I}_{\{m\}} \in \{\textbf{I}_{\{m\}}\}$}
 \STATE Extract $z$ and $\textbf{T}_{F}$ from $\textbf{I}_{\{m\}}$
\IF{$c = z \wedge \textbf{T} = \textbf{T}_{F}$}
\STATE $\{\textbf{I}_{t}\} \gets \{\textbf{I}_{t}\} \cup \textbf{I}_{\{m\}}$
\ENDIF
\ENDFOR
\ENDFOR
\FORALL{$\textbf{I}_{t} \in \{\textbf{I}_{t}\}$}
\STATE $\{\textbf{I}\} \gets \{\textbf{I}\} \cup $ findSubInstances$(\textbf{I}_{t}, \{\textbf{I}_{\{m\}}\})$
\STATE $\{\textbf{I}\} \gets \{\textbf{I}\} \cup \textbf{I}_{t}$
\ENDFOR
\RETURN{$\{\textbf{I}\}$} 
}
\end{algorithmic}
\end{algorithm}

\begin{algorithm}[H]
\begin{algorithmic}[1]
{\small
\caption{predictPose$(o_{P}, \textbf{P}_{\{m\}}^{\ast}(o_{P}), \textbf{T}_{F}) \rightarrow \textbf{T}_{P}$.}
\label{alg:asr_predictPose}
\STATE $\textbf{T}_{P} \gets \textbf{T}_{F}$
\FOR{$k \gets 1,\dots,\left|\textbf{P}_{\{m\}}^{\ast}(o_{P})\right| + 1$}
\IF{$k = \left|\textbf{P}_{\{m\}}^{\ast}(o_{P})\right| +1$}
\STATE $o \gets o_{P}$
\ELSE
\STATE $o \gets o_{F}(m_{k+1})$ with $m_{k+1}$ in $\textbf{P}_{\{m\}}^{\ast}(o_{P})$
\ENDIF
\STATE $\textbf{T}_{P} \gets$ randomVoteOnPose$(\textbf{T}_{P},m_{k}, o)$
\ENDFOR
\RETURN $\textbf{T}_{P}$
}
\end{algorithmic}
\end{algorithm}

\begin{algorithm}[H]
\begin{algorithmic}[1]
{\small
\caption{generateCloudOfPosePredictions$(\textbf{I}_{\textbf{S}},\textbf{P}_{\{m\}}^{\ast}(o),$ $n_{P}) \rightarrow \{\textbf{T}_{P}(o)\}$.}
\label{alg:asr_generateCloudOfPosePredictions}
\STATE Extract $\textbf{T}_{F}$ and $\{i\}_{\textbf{S}}$ from $\textbf{I}_{\textbf{S}}$
\FORALL{$o_{P} \in \{o\}\setminus \{i\}_{\textbf{S}}$}
\FOR{$i \gets 0,\dots,n_{P}$}
\STATE $\textbf{T}_{P} \gets$ predictPose$(o_{P}, \textbf{P}_{\{m\}}^{\ast}(o_{P}), \textbf{T}_{F})$
\STATE $\{\textbf{T}_{P}(o_{P})\} \gets \{\textbf{T}_{P}(o_{P})\} \cup \textbf{T}_{P}$
\ENDFOR
\STATE $\{\textbf{T}_{P}(o)\} \gets \{\textbf{T}_{P}(o)\} \cup \{\textbf{T}_{P}(o_{P})\}$
\ENDFOR
\RETURN $\{\textbf{T}_{P}(o)\}$
}
\end{algorithmic}
\end{algorithm}
\end{minipage}

\bibliographystyle{elsarticle-num} 
\bibliography{bibliography}





\end{document}